%% file: repicky2017adaptive.tex
\documentclass{itatnew}
\pdfoutput=1
\usepackage{amsmath,amsfonts}
\usepackage{cite}
\usepackage{calc}
\usepackage{algorithm}
\usepackage{algorithmic}
\usepackage{xstring}
\usepackage{graphicx}
\usepackage[subrefformat=parens]{subfig}
\usepackage[dvipsnames]{xcolor}
\usepackage{url}
\usepackage{wasysym}
\usepackage{enumitem}
\usepackage{textcomp}
\usepackage{tabularx}
\usepackage{booktabs}
\usepackage{xfrac}

\newcommand{\ie}{i.\,e.,\ }
\newcommand{\aka}{a.\,k.\,a.\ }
\newcommand{\iid}{i.\,i.\,d.\ }
\newcommand{\wrt}{w.\,r.\,t.\ }

\newcommand{\gm}{\ensuremath{g_m}}
\newcommand{\gmmax}{\ensuremath{g_m^\mathrm{max}}}

\newcommand{\err}{\ensuremath{\varepsilon}}
\newcommand{\kl}{\ensuremath{D_\mathrm{KL}}}
\newcommand{\trace}{\ensuremath{\mathrm{tr}}}

\newcommand{\dm}{\ensuremath{\mathrm{D}}}                     
\newcommand{\bestFED}{\ensuremath{\mathrm{\#FE}_\mathrm{t}}} 

\newcommand{\statsig}{\makebox[0pt][l]{$^\ast$}}

\newcommand{\cmaes}{\mbox{CMA-ES}}
\newcommand{\scmaes}{\mbox{S-CMA-ES}}
\newcommand{\ascmaes}{\mbox{A-S-CMA-ES}}
\newcommand{\saACMES}{$^{s*}\!$ACM-ES}

\newcommand{\bbobdatapath}{ppdata/}
\newcommand{\ppfigscaption}{Expected running time divided by dimension versus dimension}
\newcommand{\ecdfcaptiontwo}{Empirical cumulative distribution in $2\dm$}
\newcommand{\ecdfcaptionthree}{Empirical cumulative distribution in $3\dm$}
\newcommand{\ecdfcaptionfive}{Empirical cumulative distribution in $5\dm$}
\newcommand{\ecdfcaptionten}{Empirical cumulative distribution in $10\dm$}
\newcommand{\ecdfcaptiontwenty}{Empirical cumulative distribution in $20\dm$}
\newcommand{\ecdfcaptionfourty}{Empirical cumulative distribution in $40\dm$}
\newcommand{\ecdflabeltwo}{\label{fig:ECDFs02D}}
\newcommand{\ecdflabelthree}{\label{fig:ECDFs03D}}
\newcommand{\ecdflabelfour}{\label{fig:ECDFs04D}}
\newcommand{\ecdflabelfive}{\label{fig:ECDFs05D}}
\newcommand{\ecdflabelten}{\label{fig:ECDFs10D}}
\newcommand{\ecdflabeltwenty}{\label{fig:ECDFs20D}}
\newcommand{\ecdflabelfourty}{\label{fig:ECDFs40D}}
\newcommand{\ppfigslabel}{\label{fig:scaling}}

\newcommand{\aRT}{\ensuremath{\mathrm{aRT}}}

\newcommand{\Df}{\ensuremath{\Delta f}}
\newcommand{\nbFEs}{\ensuremath{\mathrm{\#FEs}}}
\newcommand{\fopt}{\ensuremath{f_\mathrm{opt}}}

\newcommand{\cocoversion}{2.1}

\begin{document}

\title{Adaptive Generation-Based Evolution Control\\for Gaussian Process Surrogate Models}

\author{Jakub Repický\inst{1,2} \and Lukáš Bajer\inst{1,2} \and Zbyněk Pitra\inst{2,3,4} \and Martin Holeňa\inst{2}}

\institute{Faculty of Mathematics and Physics, Charles University in Prague\\
Malostranské nám. 25, 118 00 Prague 1, Czech Republic\\
\and Institute of Computer Science, Czech Academy of Sciences\\
Pod Vodárenskou věží~2, 182 07 Prague 8, Czech Republic\\
\and Faculty of Nuclear Sciences and Physical Engineering, Czech Technical University in Prague\\
Břehová 7, 115 19 Prague 1, Czech Republic\\
\and National Institute of Mental Health\\
Topolová 748, 250 67 Klecany, Czech Republic\\
\email{\{repicky,bajer,pitra,martin\}@cs.cas.cz}
}

\maketitle

\section*{Errata}
The original article appeared in
Proceedings of the 17th Conference on Information Technologies---Applications and Theory (ITAT 2017) Martinské hole, Slovakia, September 22--26, 2017, published in print by CreateSpace Independent Publishing Platform and online by CEUR Workshop Proceedings.

This updated version incorporates the following changes:
\begin{description}
  \item[Abstract] Added credits to the \saACMES\ algorithm.
  \item[Section~\ref{sec:intro}] Added references and clarified the motivation.
  \item[Section~\ref{sec:ascmaes}] Added references.
\end{description}

\begin{abstract}
  The interest in accelerating black-box optimizers has resulted in several surrogate model-assisted version of the Covariance Matrix Adaptation Evolution Strategy, a state-of-the-art continuous black-box optimizer.
  The version called Surrogate CMA-ES uses Gaussian processes or random forests surrogate models with a generation-based evolution control.
  This paper presents an adaptive improvement for S-CMA-ES based on a general procedure introduced with the \saACMES\ algorithm, in which the number of generations using the surrogate model before retraining is adjusted depending on the performance of the last instance of the surrogate. 
  Three algorithms that differ in the measure of the surrogate model's performance are evaluated on the COCO/BBOB framework. 
  The results show a minor improvement on S-CMA-ES with constant model lifelengths, especially when larger lifelengths are considered.
\end{abstract}

\section{Introduction}
\label{sec:intro}
The problem of optimization of real-valued functions without a known mathematical expression, arising in many engineering tasks, is referred to as continuous black-box optimization.
Evolutionary strategies, a class of randomized population-based algorithms inspired by natural evolution, are a popular choice for continuous black-box optimization.
Especially the Covariance Matrix Adaptation Evolution Strategy (\cmaes)~\cite{hansen2010comparing} is considered the state-of-the-art continuous black-box optimizer of the several past decades.
Since values of a black-box function can only be obtained empirically and at considerable costs in practice, the number of function evaluations needed to obtain a desired function value is a key criterion for evaluating black-box optimizers.

The technique of surrogate modelling aims at saving function evaluations by building a surrogate model of the fitness and using that for a portion of function evaluations conducted in the course of the evolutionary search.
Several surrogate model-assisted versions of the \cmaes\ have been developed (see \cite{pitra2017overview} for a recent comparison of some of the most notable algorithms).
Surrogate \cmaes\ (S-\cmaes)~\cite{bajer2015benchmarking} utilizes random forests- or Gaussian processes-based surrogate models, which possess an inherent capability to quantify uncertainty of the prediction.

In order to control surrogate model's error, \scmaes\ uses the surrogate model for a given number of generations $g_m$ before a new instance of the model is trained on a population evaluated with the fitness, which is a strategy called generation-based evolution control~\cite{jin2002fitness}.
In~\cite{bajer2015benchmarking}, two values, in particular $g_m \in \{1,5\}$, have been benchmarked on the COCO/BBOB framework.
In many cases, the higher value of $g_m$ outperformed the lower one in earlier phases of the optimization, but the reverse order was observed towards later phases of the optimization.

The \saACMES\ algorithm~\cite{loshchilov2012self} introduced an adaptive evolution control adjusting surrogate hyperparameters and \emph{lifelength}, \ie the number of model-evaluated generations, as a function of previous model's error.

In this paper, we use the procedure for adjusting $g_m$ from \saACMES\ in connection with three different surrogate model error measures.
The three \scmaes\ versions are compared on the COCO/BBOB framework.
We restrict our attention to \scmaes\ with Gaussian processes, since they outperformed random forest-based surrogates~\cite{bajer2015benchmarking}.

The remainder of this paper is organized as follows.
Section~\ref{sec:scmaes} outlines basic concepts of \scmaes.
The adaptive version is described~in~Section~\ref{sec:ascmaes}.
Experimental setup is given in~Section~\ref{sec:expsetup}.
Experimental results are reported in~Section~\ref{sec:results}.
Section~\ref{sec:conclusion} concludes the paper.

\section{Surrogate \cmaes}
\label{sec:scmaes}
The \cmaes\ operates on a population of $\lambda$ candidate solutions sampled from a multivariate normal distribution:
\begin{equation}
  \mathbf{x}_k \sim \mathcal{N}(\mathbf{m}, \sigma^2\mathbf{C})\quad k = 1,\dots,\lambda,
  \label{eq:cmaes:sampling}
\end{equation}
where $\mathcal{N}$ is the normal distribution function; $\mathbf{m}$ and $\mathbf{C}$ are the mean and the covariance matrix of the estimated search distribution, respectively; and the $\sigma$ is the overall search step size.
The candidate solutions are ranked according to their fitness values:
\begin{equation}
  y_k = f(\mathbf{x}_k)\quad k = 1,\dots,\lambda.
  \label{eq:cmaes:eval}
\end{equation}
Upon a (weighted) selection of $\mu < \lambda$ highest ranked points, the mean and the covariance matrix of the multivariate normal distribution are adapted according to a procedure that takes as input, among other variables, a cumulation of the past search steps \cite{hansen2016cma}. 
The \scmaes\ modifies the \cmaes\ by replacing its sampling~\eqref{eq:cmaes:sampling} and fitness-evaluation~\eqref{eq:cmaes:eval} steps with a procedure depicted in~Algorithm~\ref{alg:scmaes}.

\begin{algorithm}
  \begin{algorithmic}[1]
    \REQUIRE{$g$ (generation)\\
    \gm\ (number of model-evaluated generations)\\
    $\sigma, \lambda, \mathbf{m}, \mathbf{C}$ (\cmaes\ internal variables)\\
    $r$ (maximal distance between $\mathbf{m}$ and a training point)\\
    $n_\text{req}$ (minimal number of points for training)\\
    $n_\text{max}$ (maximal number of points for training)\\
    $\mathcal{A}$ (archive), $f_\mathcal{M}$ (model), $f$ (fitness)}
    \STATE $\mathbf{x}_k \sim \mathcal{N}(\mathbf{m}, \sigma^2\mathbf{C})\quad k = 1,\dots,\lambda$ \COMMENT{sampling} \label{step:scmaes:sampling}
    \IF{$g$ is original-fitness-evaluated} \label{step:scmaes:switch}
      \STATE $y_k \leftarrow f(\mathbf{x}_k)\quad k = 1,\dots,\lambda$ \COMMENT{fitness evaluation} \label{step:scmaes:feval}
    \STATE $\mathcal{A} \leftarrow \mathcal{A} \cup \left\{\left(\mathbf{x}_k, y_k \right)\right\}_{k = 1}^\lambda$ \label{step:scmaes:updatearch}
    \STATE{$\left(\mathbf{X}_\mathrm{tr}, \mathbf{y}_\mathrm{tr}\right) \leftarrow $ choose $n_\text{tr}$ training points within the Mahalanobis distance $r$ from $\mathcal{A}$, assuring that $n_\text{req} \le n_\text{tr} \le n_\text{max}$}\label{step:scmaes:trselect}
    \STATE $f_\mathcal{M} \leftarrow \text{train\_model}(\mathbf{X}_\mathrm{tr}, \mathbf{y}_\mathrm{tr})$ \label{step:scmaes:training}
    \STATE mark $(g + 1)$ as model-evaluated
    \ELSE
    \STATE $\hat{y}_k \leftarrow f_\mathcal{M}(\mathbf{x}_k)$ \COMMENT{model evaluation} \label{step:scmaes:modeval}
    \IF{\gm\ model generations have passed}
    \STATE mark $(g + 1)$ as original-fitness-evaluated
    \ENDIF
    \ENDIF
    \ENSURE{$f_\mathcal{M},\:\mathcal{A},\:(y_k)_{k=1}^\lambda$}
  \end{algorithmic}
  \caption{Surrogate part of \scmaes}
  \label{alg:scmaes}
\end{algorithm}

Depending on the generation number $g$, the procedure evaluates all candidate solutions either with the real fitness or with the model.
In each case, the sampling of the estimated multivariate normal distribution is unchanged (step~\ref{step:scmaes:sampling}).

If the population is original-fitness-evaluated (step~\ref{step:scmaes:feval}), the new evaluations are saved in an archive of known solutions (step~\ref{step:scmaes:updatearch}).
Afterwards, a new model is trained on a set of points within the Mahalanobis distance $r$ from the current \cmaes\ distribution $\mathcal{N}(\mathbf{m}, \sigma\mathbf{C})$ (step~\ref{step:scmaes:trselect}).

In model-evaluated generations,
the fitness values of the whole population of candidate solutions are estimated by the model (step~\ref{step:scmaes:modeval}).

\subsection{Gaussian Processes}
A Gaussian process (GP) is a collection of random variables $(f(\mathbf{x}))_{\mathbf{x}\in\mathbb{R}^D}$, such that any finite subcollection $\mathbf{f} = (f(\mathbf{x}_1),\dots,f(\mathbf{x}_N))$ has an $N$-dimensional normal distribution.
A Gaussian process is defined by a mean function $\mu(\mathbf{x})$ (often assumed to be zero) and a covariance function $k(\mathbf{x}, \mathbf{x}; \mathbf{\theta})$, where $\mathbf{\theta}$ is a vector of parameters of $k$, hence hyperparameters of the Gaussian process.
Given a set of training data $X = \{\mathbf{x}_1,\dots,\mathbf{x}_N\}$, the covariance matrix of a GP prior is $\mathbf{K}_N + \sigma_n^2\mathbf{I}_N$, where $\mathbf{K}_N$ is a $N \times N$ matrix given by $\{\mathbf{K}_N\}_{i,j} = k(\mathbf{x}_i, \mathbf{x}_j; \mathbf{\theta})$ for all $i, j = 1,\dots,N$; $\sigma_n^2$ is the variance of an additive, \iid noise and $\mathbf{I}_N$ is a $N \times N$ identity matrix.
Given a new point $\mathbf{x}_\ast \notin X$, Gaussian process regression is derived by conditioning the joint normal distribution of $(f(\mathbf{x}_1),\dots,f(\mathbf{x}_N),f(\mathbf{x}_\ast))$ on the prior, which yields a univariate Gaussian (see \cite{rassmusen2006gaussian} for more details).
The hyperparameters $\mathbf{\theta}$ of a GP regression model are estimated using the maximum likelihood estimation method.

\section{Adaptive Evolution Control for Surrogate CMA-ES}
\label{sec:ascmaes}

The generation-based evolution strategy optimizes the fitness function and the surrogate model thereof in certain proportion.
On problem areas that can be approximated well, a surrogate-assisted optimization might benefit from frequent utilization of the model, while on areas that are hard for the surrogate to approximate, frequent utilization of the model might degenerate the performance due to the model's inaccuracy.

Adaptation of the number of model evaluated generations $g_m$ (in addition to other surrogate model parameters that we don't investigate here) depending on the previous model's error has been proposed in~\saACMES~\cite{loshchilov2012self}.

Let $g$ be a generation that is marked as original-fitness-evaluated, and a newly-trained surrogate model $f_\mathcal{M}$.
If $f_\mathcal{M}$ is the first surrogate trained so far, put $g_m~=~1$.
Otherwise, an error $\err$ of a previous surrogate model $f_\mathcal{M}^\mathrm{last}$ is estimated on the newly evaluated population $(\mathbf{x}_1^{(g+1)},\dots,\mathbf{x}_\lambda^{(g+1)})$ (Algorithm~\ref{alg:ascmaes:estimate}).
The error $\err$ is then mapped into
a number of consecutive generations $\gm,\,\gm \in [0,\gmmax]$, for which the surrogate $f_\mathcal{M}$ will be used (Algorithm~\ref{alg:ascmaes:update}).

We investigate three approaches for expressing surrogate model error.
As the \cmaes\ depends primarily on the ranking of candidate solutions, the first two approaches, \emph{Kendall correlation coefficient} and \emph{Rank difference} are based on ranking.
The third one, previously proposed in~\cite{loshchilov2013kl}, uses \emph{Kullback-Leibler divergence} \aka \emph{information gain} to measure a difference between a multivariate normal distribution estimated from the fitness values~$\mathbf{y}$ and a multivariate normal distribution estimated for the predicted values~$\hat{\mathbf{y}}$.

\begin{algorithm}
  \begin{algorithmic}[1]
    \REQUIRE{error\_type (one of \{``Kendall'', ``Rank-Difference'', ``Kullback-Leibler''\})\\
    $g$ (\cmaes\ generation number)\\
    $\mathbf{x}_1^{(g+1)},\dots,\mathbf{x}_\lambda^{(g+1)}$ (a newly sampled population)\\
    $\mathbf{y}, \hat{\mathbf{y}}$ (fitness values and model predictions in generation $g$)\\
    $\mathbf{c}_\mathrm{cma} = (c_c, c_1, c_\mu, c_\sigma, d_\sigma)$ (\cmaes\ constants)\\
    $\mathbf{v}_\mathrm{cma}^{(g)} = (\mathbf{m}^{(g)}, \mathbf{C}^{(g)}, \mathbf{p}_\sigma^{(g)}, \mathbf{p}_c^{(g)}, \sigma^{(g)})$ (\cmaes\ variables at generation~$g$)\\
    $\err_\mathrm{max}$ (maximal error so far)}
    \IF{error\_type = ``Kendall''}
      \STATE $\tau \leftarrow $ Kendall rank correlation coefficient between $\mathbf{y}$ and $\hat{\mathbf{y}}$
      \STATE $\err \leftarrow \frac{1}{2}(1-\tau)$
    \ELSIF{error\_type = ``Rank-Difference''}
      \STATE $\err \leftarrow \err_\mathrm{RD}^\mu(\hat{\mathbf{y}}, \mathbf{y})$
    \ELSIF{error\_type = ``Kullback-Leibler''}
     \STATE $(\mathbf{m}^{(g+1)}, \mathbf{C}^{(g+1)}, \sigma^{(g+1)}) \leftarrow\newline
     \mathrm{cma\_update}((\mathbf{x}_1^{(g+1)},\dots,\mathbf{x}_\lambda^{(g+1)}), \mathbf{y}, \mathbf{c}_\mathrm{cma}, \mathbf{v}_\mathrm{cma}^{(g)})$ \label{step:ascmaes:estimate:cma-a}
     \STATE $(\mathbf{m}_\mathcal{M}^{(g+1)}, \mathbf{C}_\mathcal{M}^{(g+1)}, \sigma^{(g+1)}) \leftarrow\newline
     \mathrm{cma\_update}((\mathbf{x}_1^{(g+1)},\dots,\mathbf{x}_\lambda^{(g+1)}), \hat{\mathbf{y}},\mathbf{c}_\mathrm{cma}, \mathbf{v}_\mathrm{cma}^{(g)})$ \label{step:ascmaes:estimate:cma-b}
     \STATE $\err \leftarrow \kl(\mathcal{N}(\mathbf{m}_\mathcal{M}^{(g+1)}, \sigma_\mathcal{M}^{(g+1)}\mathbf{C}_\mathcal{M}^{(g+1)})\|\newline
     \mathcal{N}(\mathbf{m}^{(g+1)}, \sigma^{(g+1)}\mathbf{C}^{(g+1)}))$
     \IF{$\err > \err_\mathrm{max}$}
       \STATE $\err_\mathrm{max} \leftarrow \err$
     \ENDIF
     \STATE $\err \leftarrow \frac{\err}{\err_\mathrm{max}}$ \COMMENT{normalize in proportion to the historical maximum}\label{step:ascmaes:estimate:norm}
    \ENDIF
    \ENSURE{$\err \in [0, 1]$}
  \end{algorithmic}
  \caption{Model error estimation}
  \label{alg:ascmaes:estimate}
\end{algorithm}

\paragraph{Kendall rank correlation coefficient}
Kendall rank correlation coefficient $\tau$ measures similarity between two different orderings of the same set.
Let $\mathbf{y} = \left(f\left(\mathbf{x}_1\right),\dots,f\left(\mathbf{x}_\lambda\right)\right)$ and $\hat{\mathbf{y}} = \left(f_\mathcal{M}^\mathrm{last}(\mathbf{x}_1),\dots,f_\mathcal{M}^\mathrm{last}(\mathbf{x}_\lambda)\right)$ be the sequences of the fitness values and the predicted values of a population $\mathbf{x}_1,\dots,\mathbf{x}_\lambda$, respectively.
A pair of indices $(i, j)$, such that $i \neq j,\,i,j \in \{1,\dots,\lambda\}$, is said to be concordant, if both $y_i < y_j$ and $\hat{y}_i < \hat{y}_j$ or if both $y_i > y_j$ and $\hat{y}_i > \hat{y}_j$.
A discordant pair $(i, j), i\neq j,\,i,j \in \{1,\dots,\lambda\}$ is one fulfilling that both $y_i < y_j$ and $\hat{y}_i > \hat{y}_j$ or both $y_i > y_j$ and $\hat{y}_i < \hat{y}_j$.
Let $n_c$ and $d_c$ denote the number of concordant and discordant pairs of indices from $\{1,\dots,\lambda\}$, respectively.
The Kendall correlation coefficient $\tau$ between vectors $\mathbf{y}$ and $\hat{\mathbf{y}}$ is defined as:
\begin{equation*}
  \label{eq:kendall}
  \tau = \frac{2}{\lambda(\lambda-1)} (n_c - n_d).
\end{equation*}
In the corresponding branch of Algorithm~\ref{alg:ascmaes:estimate}, the value $\tau$ is decreasingly scaled into interval $[0, 1]$.

\paragraph{Ranking difference error}
The ranking difference error is a normalized version of a measure used in~\cite{kern2006local}.
Given $r_1(i)$ the rank of the $i$-th element of $\hat{\mathbf{y}}$ and $r_2(i)$ the rank of the $i$-th element of $\mathbf{y}$, the ranking difference error is the sum of element-wise differences between $r_1$ and $r_2$ taking into account only the $\mu$ best-ranked points from~$\hat{\mathbf{y}}$:
\begin{equation*}
  \err^\mu_\mathrm{RD}(\hat{\mathbf{y}}, \mathbf{y}) = \frac{ \sum_{i: r_1(i) \leq \mu} \left| r_2(i) - r_1(i) \right| } { \max_{\pi \, \in \, S_\lambda} \sum_{i: \pi(i) \leq \mu} | \, i - \pi(i) | },
  \label{eq:rde}
\end{equation*}
where $S_\lambda$ is the group of all permutations of set $\{1,\dots,\lambda\}$.

\paragraph{Kullback-Leibler divergence}
Kullback-Leibler divergence from a continuous random variable $Q$ with probability density function $q$ to a continuous random variable $P$ with probability density function $p$ is defined as:
\begin{equation*}
  \label{eq:kl}
  \kl(P \| Q) = \int_{-\infty}^\infty\, p(x)\,\log\,\frac{p(x)}{q(x)}\,dx.
\end{equation*}
For two multivariate normal distributions $\mathcal{N}_1(\mu_1, \Sigma_1)$ and $\mathcal{N}_2(\mu_2,\Sigma_2)$ with the same dimension $k$, the Kullback-Leibler divergence from $\mathcal{N}_2$ to $\mathcal{N}_1$ is:
\begin{alignat*}{2}
  \kl(\mathcal{N}_1\|\mathcal{N}_2) = &\frac{1}{2}\Bigg(\trace({\Sigma}_2^{-1}{\Sigma}_1) + \ln\left(\frac{\left|{\Sigma}_2\right|}{\left|{\Sigma}_1\right|}\right) +\\
  &(\mu_2 - \mu_1)^T {\Sigma}_2^{-1}(\mu_2 - \mu_1) - k\Bigg).
\end{alignat*}

The algorithm of model error estimation (Algorithm~\ref{alg:ascmaes:estimate}) in generation $g$ computes Kullback-Leibler divergence from a CMA-estimated multivariate normal distribution $\mathcal{N}(\mathbf{m}^{(g+1)}, \mathbf{C}^{(g+1)})$ \wrt fitness values $\mathbf{y}$ to a CMA-estimated multivariate normal distribution $\mathcal{N}(\mathbf{m}_\mathcal{M}^{(g+1)}, \mathbf{C}_\mathcal{M}^{(g+1)})$ \wrt predicted values $\hat{\mathbf{y}}$. 
Procedure cma\_update in steps~\ref{step:ascmaes:estimate:cma-a} and \ref{step:ascmaes:estimate:cma-b} refers to one iteration of the \cmaes\ from the point when a new population has been sampled.
The result is normalized by the historical maximum (step~\ref{step:ascmaes:estimate:norm}).

\begin{figure}
  \centering
  \includegraphics[width=0.35\textwidth]{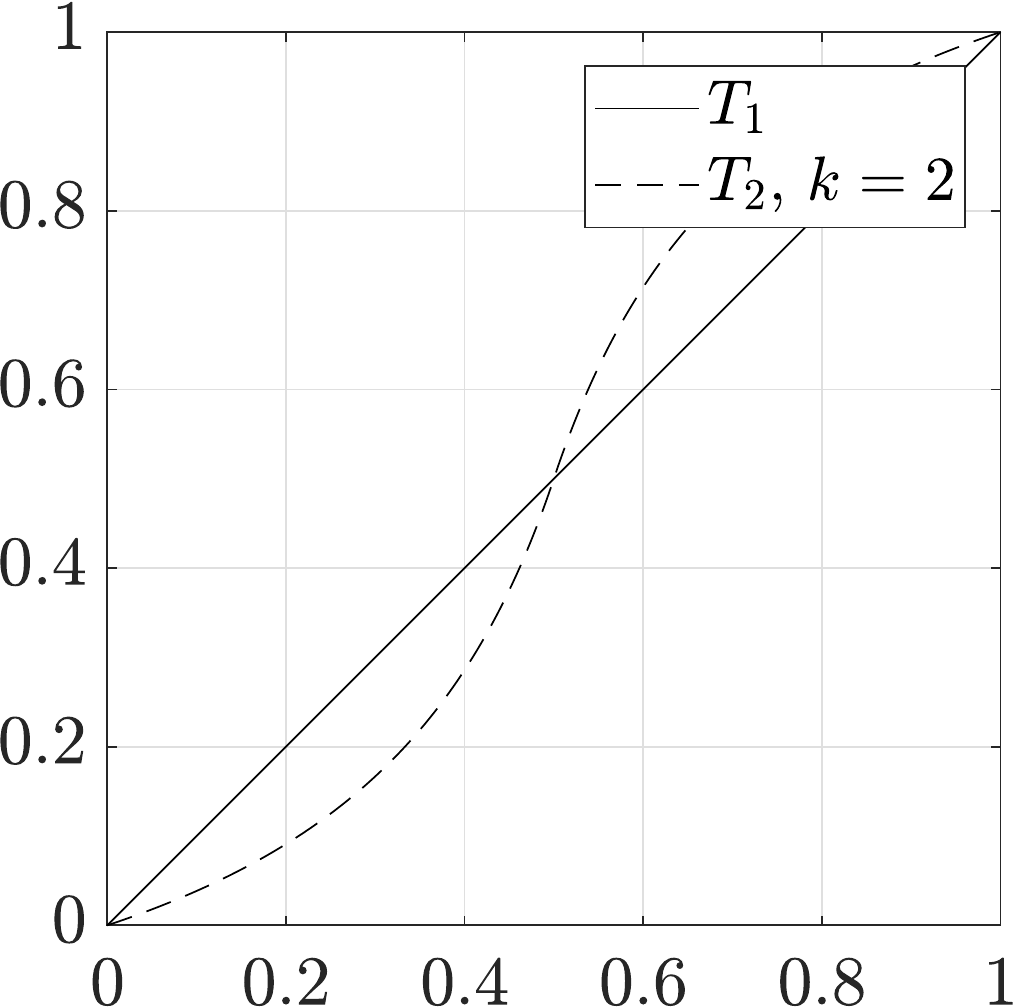}
  \caption{Model error transfer functions}
  \label{fig:transfer}
\end{figure}

\begin{algorithm}
  \begin{algorithmic}[1]
    \REQUIRE{\err\ (estimation of surrogate model error, $\err \in [0, 1]$)\\
      $\err_T \in [0, 1]$ (a threshold at which the error is truncated to 1)\\
    $\gamma\colon [0, 1] \to [0, 1]$ (transfer function)\\
    $r_u$ (error update rate)\\
    $\err_\mathrm{last}$ (model error from the previous iteration)\\
    $\gmmax$ (upper bound for admissible number of model generations)\\
    }
    \STATE $\err \leftarrow (1 - r_u)\err_\mathrm{last} + r_u\err$ \COMMENT{exponential smoothing} \label{step:ascmaes:update:smoothing}
    \STATE $\err_\mathrm{last} \leftarrow \err$
    \STATE $\err \leftarrow \frac{1}{\err_T}\min{\{\err, \err_T\}}$ \COMMENT{truncation to $1$} \label{step:ascmaes:update:trunc}
    \STATE $\gm \leftarrow \mathrm{round}(\gamma(1-\err)\gmmax) $ \COMMENT{scaling into the admissible interval} \label{step:ascmaes:update:transf}
    \ENSURE{\gm\ -- updated number of model-evaluated generations}
  \end{algorithmic}
  \caption{Updating the number of model generations}
  \label{alg:ascmaes:update}
\end{algorithm}

\paragraph{Adjusting the number of model generations}
The model of dependence of the number of consecutive model generations $g_m$ on the model error (Algorithm~\ref{alg:ascmaes:update}) is almost identical to the approach used in~\cite{loshchilov2012self}.
The history of surrogate model errors $\err$ is exponentially smoothed with a rate $r_u$ (step~\ref{step:ascmaes:update:smoothing}).
The error is truncated at a threshold $\err_T$ so that resulting $\gm = \gmmax$ for all values $\err \geq \err_T$ (step~\ref{step:ascmaes:update:trunc}).
In contrast to~\cite{loshchilov2012self}, we consider two different transfer functions $T_1,T_2\colon [0,1] \to [0,1]$ (plotted in~Figure~\ref{fig:transfer}) that scale the error into the admissible interval $[0, \gmmax]$:
\begin{alignat}{2}
  T_1(x) &= x \label{eq:transf:id}\\
  T_2(x; k) &= \frac{\left(x-\frac{1}{2}\right)\left(1 + \frac{1}{k}\right)}{
  \left|2\left(x-\frac{1}{2}\right)\right| + \frac{1}{k}} + \frac{1}{2},\, k > 0 \label{eq:transf:sig:a}.
\end{alignat}
Both functions are defined on $[0, 1]$, moreover, $T_i(0) = 0$ and $T_i(1) = 1$ for $i=1,2$.
Transfer function $T_2$ is a simple sigmoid function defined to be slightly less sensitive near the edges than in the middle.
More control can thus be achieved in the regions of low and high error values.
The parameter $k$ determines the steepness of the sigmoid curve.

\section{Experimental Setup}
\label{sec:expsetup}
The proposed adaptive generation-based evolution control for the \scmaes\ with three different surrogate model error measures is evaluated on the noiseless testbed of the COCO/BBOB (Comparing Continuous Optimizers / Black-Box Optimization Benchmarking) framework \cite{hansen2009noiseless,hansen2012experimental} and compared with the \scmaes\ and \cmaes.

Each function is defined everywhere on $\mathbb{R}^\dm$ and has its optimum in $[-5, 5]^\dm$ for all dimensionalities $D \geq 2$.
For every function and every dimensionality, $15$ trials of the optimizer are run on independent instances, which differ in linear transformations of the $x$-space or shifts of the $f$-space.
In our experiments, instances recommended for BBOB 2015 workshop, \ie $\{1,\dots,5,41,\dots50\}$, were used.
Each trial is terminated when the $\fopt$ is reached within a small tolerance $\Df_t = 10^{-8}$ or when a given budget of function evaluations, $250\dm$ in our case, is used up.
Experiments were run for dimensionalities $2$, $3$, $5$, $10$ and~$20$.
The algorithms' settings are summarized in the following subsections.

\subsection{\cmaes}
The \cmaes\ results in BBOB format were downloaded from the BBOB 2010 workshop archive~\footnote{\url{http://coco.gforge.inria.fr/data-archive/bbob/2010/}}.
The \cmaes\ used in those experiments was in version 3.40.beta
and utilized a restart strategy (known as IPOP-\cmaes), where the population size is increased by factor IncPopSize after each restart~\cite{auger2005restart}.
The default parameter values employed in the \cmaes\ are $\lambda = 4 + \lfloor 3\text{log}D\rfloor$, $\mu = \lfloor \frac{\lambda}{2} \rfloor$, $\sigma_{\text{start}} = \frac{8}{3}$, $\text{IncPopSize} = 2$.

\subsection{\scmaes}
The \scmaes\ was tested with two numbers of model-evaluated generations, $\gm = 1$ (further denoted as ``GP-1'') and $\gm = 5$ (``GP-5'').
All other \scmaes\ settings were left as described in~\cite{bajer2015benchmarking}.
In particular, the Mahalanobis distance was $r = 8$, the starting values $(\theta, l)$ of the Mat\'{e}rn covariance function $k^{\nu=5/2}_\text{Mat\'{e}rn}$ were $(0.5, 2)$ and the starting value of the GP noise parameter $\sigma_n^2$ was $0.01$.
If not mentioned otherwise, the corresponding settings of adaptive versions of the \scmaes\ are as just stated.

\begin{table}
  \centering
  \caption{Discretization of the \ascmaes\ parameters.}
  \begin{tabular}{ll}
    \toprule
    \textbf{Parameter} & \textbf{Discretization}\\
    \midrule
    $\gamma$ & $T_1$~\eqref{eq:transf:id}, $T_2$~\eqref{eq:transf:sig:a}\\
    $\err_T$ & $0.5, 0.9$\\
    $\gm$ & $5, 10, 20$\\
    $r_u$ & $0.2, 0.5, 0.8$\\
    \bottomrule
  \end{tabular}
  \label{tab:factors}
\end{table}

In order to find the most promising settings for each considered surrogate error measure, a full factorial experiment was conducted on one half of the noiseless testbed, namely on functions $f_i$ for $i \in \{2, 3, 6, 8, 12, 13, 15, 17, 18, 21, 23, 24\}$.
The discretization of continuous parameters $(\gamma, \err_T, \gmmax, r_u)$ is reported in~Table~\ref{tab:factors}.
All possible combinations of the parameters
were ranked on the $12$ selected functions according to the lowest achieved $\Df^\text{med}$ (see~Section~\ref{sec:results}) for different numbers of function evaluations $\nbFEs/D = 25, 50, 125, 250$.
The best settings were chosen according to the highest sum of $1$-st rank counts.
Ties were resolved according to the lowest sum of ranks.
All of the best settings included maximum model-evaluated generations $\gmmax = 5$.
The remaining of the winning values are summarized in the following paragraphs.

\paragraph{Kendall correlation coefficient (ADA-Kendall)}
Transfer function $\gamma = T_2$, error threshold $\err_T = 0.5$ and update rate $r_u = 0.2$.
\paragraph{Ranking difference error (ADA-RD)}
The same, except transfer function was $\gamma = T_1$.
\paragraph{Kullback-Leibler divergence (ADA-KL)}
Transfer function $\gamma = T_2$, error threshold $\err_T = 0.9$ and update rate $r_u = 0.5$.

\section{CPU Timing}
\input{tables/adaTiming}
In order to assess computational costs other than the number of function evaluations, we calculate CPU timing per function evaluation for each algorithm and each dimensionality.
Each experiment was divided into jobs by dimensionalities, functions and instances.
All jobs were run in a single thread on the Czech national grid MetaCentrum.
The average time per function evaluation for each algorithm and each tested dimensionality is summarized in~Table~\ref{tab:timing:ada}.

\section{Results}
\label{sec:results}

\newlength{\dueltabcolw}
\newlength{\savetabcolsep}
\newlength{\savecmidrulekern}
\newlength{\firstcolw}

\newlength{\astwidth}
\newcommand{\statstabcap}{%
  \caption[Mean ranks of Adaptive S-CMA-ES versions]{Mean ranks of the \cmaes, the \scmaes\ and all \ascmaes\ versions over the BBOB and the Iman-Davenport variant of the Friedman test for the 10 considered combinations of dimensionalities and evaluation budgets. The lowest value is highlighted in bold. Statistically significant results at the significance level $\alpha = 0.05$ are marked by an asterisk.}}
\newcommand{\statstablab}{\label{tab:stats:ada}}
\input{tables/adaStatsTable2}

We test the difference in algorithms' convergence for significance on the whole noiseless testbed with the non-para\-metric Friedman test~\cite{demsar2006statistical}.
The algorithms are ranked on each BBOB function with respect to medians of log-scaled minimal distance $\Df$ from the function optimum, denoted as $\Df^\text{med}$, at a fixed budget of function evaluations.

To account for different optimization scenarios,
the test is conducted separately for all considered dimensionalities of the input space and two function evaluation budgets, a higher and a lower one.
Let \bestFED\ be the smallest number of function evaluations at which at least one algorithm reached the target, \ie satisfied $\Df^\text{med} \le \Df_t$,
or $\bestFED = 250\dm$ if the target has not been reached.
We set the higher budget for the tests to \bestFED\ and the lower budget to $\frac{\bestFED}{3}$.

Mean ranks from the Friedman test are given~in~Table~\ref{tab:stats:ada}.
The critical value for the Friedman test is $2.29$.

The mean ranks differ significantly for all tested scenarios except for both tested numbers of function evaluations in $2\dm$ and the higher tested number of function evaluations in $3\dm$.
Starting from $5\dm$ upwards, the lowest mean rank is achieved either by ADA-Kendall or ADA-RD at both tested $\nbFEs$.

In order to show pairwise differences, we perform a pairwise $N \times N$ comparison of the algorithms' average ranks by the post-hoc Friedman test with the Bergmann-Hommel correction of the family-wise error~\cite{garcia2008extension} in cases when the null hypothesis of equal algorithms' performance was rejected.
To better illustrate algorithms differences, we also count the number of benchmark functions at which one algorithm achieved a higher rank than the other.
The pairwise score and the statistical significance of the pairwise mean rank differences are reported in Table~\ref{tab:duel:ada}.
In the post-hoc test, ADA-Kendall significantly outperforms both the \cmaes\ and GP-5 in $10\dm$ and $20\dm$.
It also significantly outperforms GP-1 in $10\dm$ at the higher tested \nbFEs.  

\begin{figure}
  \centering
  \includegraphics[width=0.5\textwidth]{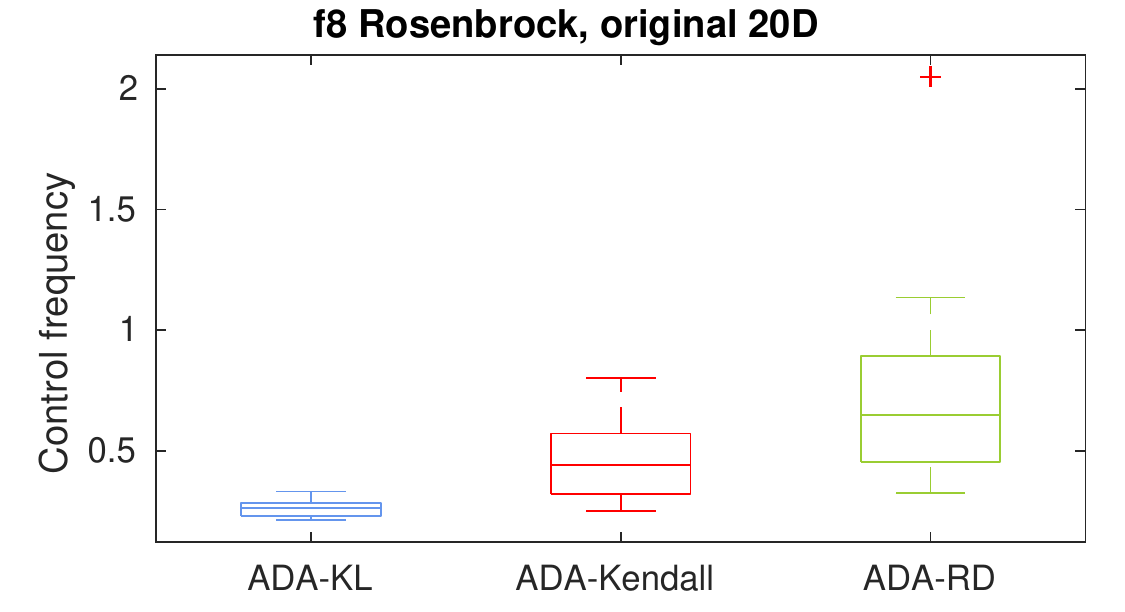}
  \caption{Average control frequency (the ratio of the number of total original-fitness-evaluated generations to the number of total model-evaluated generations) in \ascmaes\ measured in 15 trials of each algorithm on $f_8$ in $20\dm$.}
  \label{fig:boxplots}
\end{figure}

For illustration, the average control frequency given by the ratio of the number of total original-fitness-evaluated generations to the number of total model-evaluated generations within one trial, for data from $15$ trials on $f_8$ (Rosenbrock's function) in $20\dm$ is given in~Figure~\ref{fig:boxplots}.
The algorithm \mbox{ADA-KL} led to generally lower control frequencies than its competitors, which might explain its slightly inferior performance.
Similar results were observed for the remaining functions and dimensionalities.

The cases when ADA-Kendall and ADA-RD are able to switch between more exploitation-oriented and more data-gathering-oriented behaviour can be studied on the results from COCO's postprocessing.
GP-5 outperforms both GP-1 and the \cmaes\ on the lower and middle parts of the empirical distribution functions (ECDFs) basically for all dimensionalities (Figure~\ref{fig:ECDFs20D:ada}).
On the other hand, GP-1 outperforms GP-5 especially in later phases of the search (Figure~\ref{fig:ECDFs20D:ada}).

The ability of ADA-Kendall and ADA-RD to switch to a less-exploitation mode when appropriate is eminent on the ECDFs plots
in $20\dm$, especially on the moderate and the all-function groups (top right and bottom right on Figure~\ref{fig:ECDFs20D:ada}),
with exception of the well structured multimodal group (middle right), when they fail in the middle part and the weakly structured multimodal group (bottom left), when they fail towards the end of the search.

\setlength{\firstcolw}{2cm}
\newcommand{\dueltabcap}{%
  \caption[A pairwise comparison of Adaptive S-CMA-ES versions]{A pairwise comparison of the algorithms in $2\dm$, $3\dm$, $5\dm$, $10\dm$ and $20\dm$ over the BBOB for 2 different evaluation budgets.
The comparison is based on medians over runs on 15 instances for  each of all the 24~functions.
The number of wins of $i$-th algorithm against $j$-th algorithm over all benchmark functions is given in $i$-th row and $j$-th column.
The asterisk marks the row algorithm achieving a significantly lower value of the objective function than the column algorithm according to the Friedman post-hoc test with the Bergmann-Hommel correction at family-wise significance level $\alpha=0.05$.}}
\newcommand{\dueltablab}{\label{tab:duel:ada}}
\input{tables/adaDuelTable2}

\renewcommand{\bbobdatapath}{ppdata/}
\input{\bbobdatapath cocopp_commands.tex}
\renewcommand{\ppfigscaption}{Average running time of Adaptive \scmaes\ versus dimension}
\renewcommand{\ppfigslabel}{\label{fig:scaling:ada}}
\newcommand{\ecdfada}[1]{ECDFs for Adaptive \scmaes\ in $#1\dm$}
\renewcommand{\ecdfcaptiontwo}{\ecdfada{2}}
\renewcommand{\ecdfcaptionthree}{\ecdfada{3}}
\renewcommand{\ecdfcaptionfive}{\ecdfada{5}}

\renewcommand{\ecdfcaptionten}{\ecdfada{10}}
\renewcommand{\ecdfcaptiontwenty}{\ecdfada{20}}
\renewcommand{\ecdfcaptionfourty}{\ecdfada{40}}
\renewcommand{\ecdflabeltwo}{\label{fig:ECDFs02D:ada}}
\renewcommand{\ecdflabelthree}{\label{fig:ECDFs03D:ada}}
\renewcommand{\ecdflabelfour}{\label{fig:ECDFs04D:ada}}
\renewcommand{\ecdflabelfive}{\label{fig:ECDFs05D:ada}}
\renewcommand{\ecdflabelten}{\label{fig:ECDFs10D:ada}}
\renewcommand{\ecdflabeltwenty}{\label{fig:ECDFs20D:ada}}
\renewcommand{\ecdflabelfourty}{\label{fig:ECDFs40D:ada}}

\providecommand{\rot}{}
\renewcommand{\rot}[2][2.5]{
  \hspace*{-3.5\baselineskip}%
  \begin{rotate}{90}\hspace{#1em}#2
  \end{rotate}}
\providecommand{\includeperfprof}{}
\renewcommand{\includeperfprof}[1]{
  \includegraphics[width=0.38\textwidth,trim=0mm 0mm 0mm 10mm, clip]{ppdata/#1}%
}

\small


\begin{figure*}
   \small
   \centering
 \begin{tabular}{@{}c@{}c@{}}
 separable functions & moderate functions \\
 \includegraphics[width=0.38\textwidth,trim=0mm 0mm 0mm 10mm, clip]{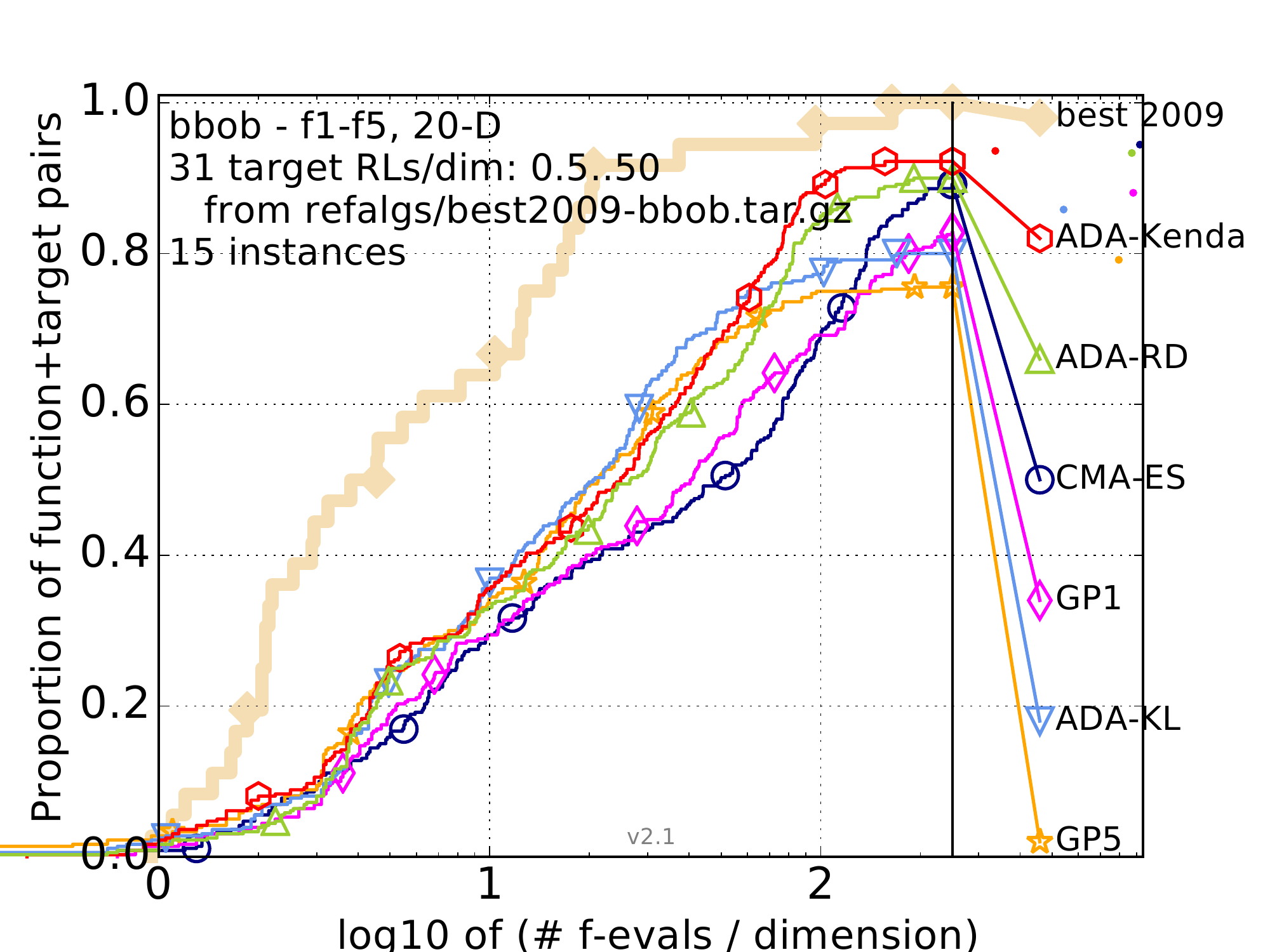} &
 \includegraphics[width=0.38\textwidth,trim=0mm 0mm 0mm 10mm, clip]{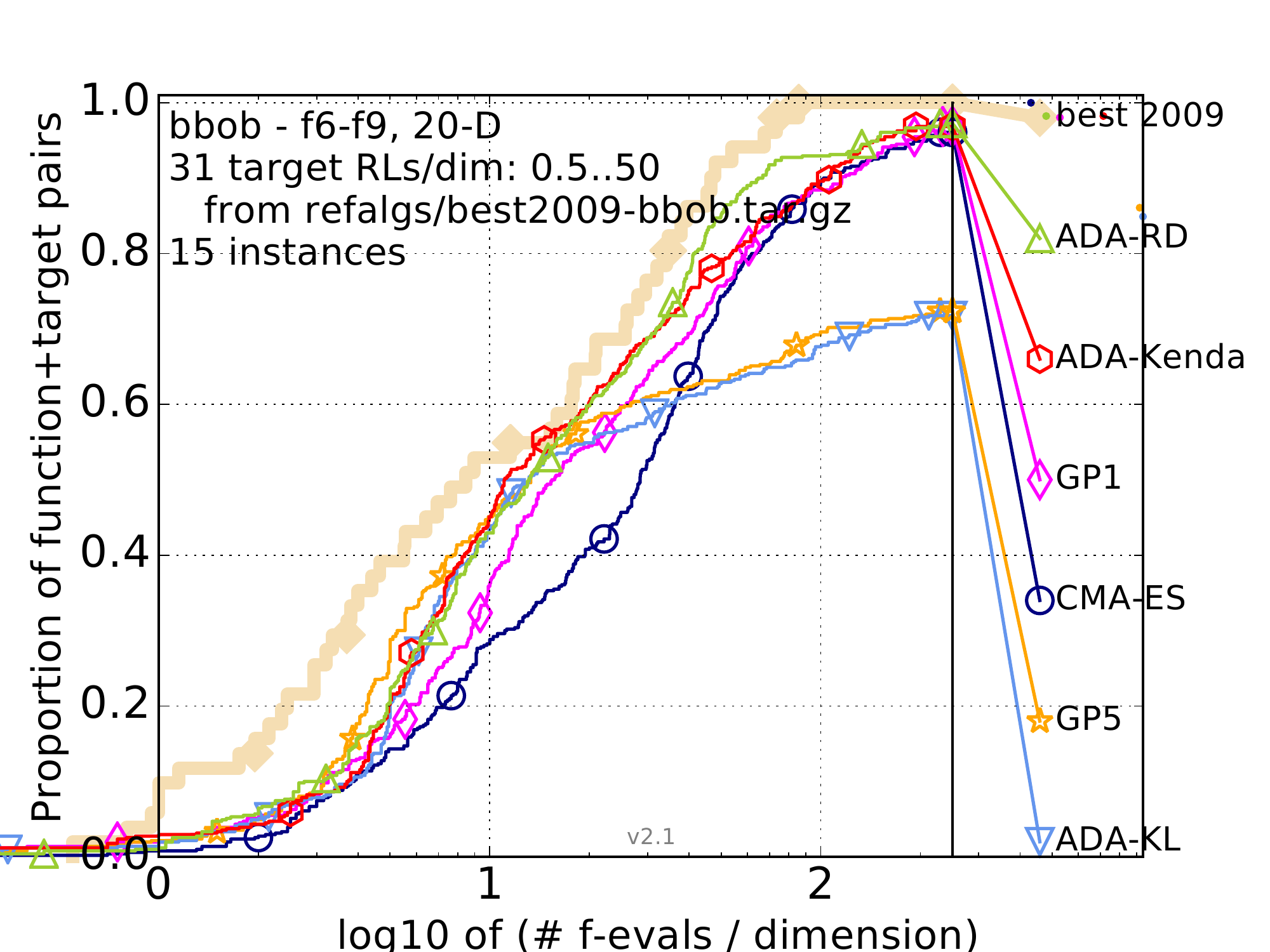} \\ 
ill-conditioned functions & multimodal functions \\
 \includegraphics[width=0.38\textwidth,trim=0mm 0mm 0mm 10mm, clip]{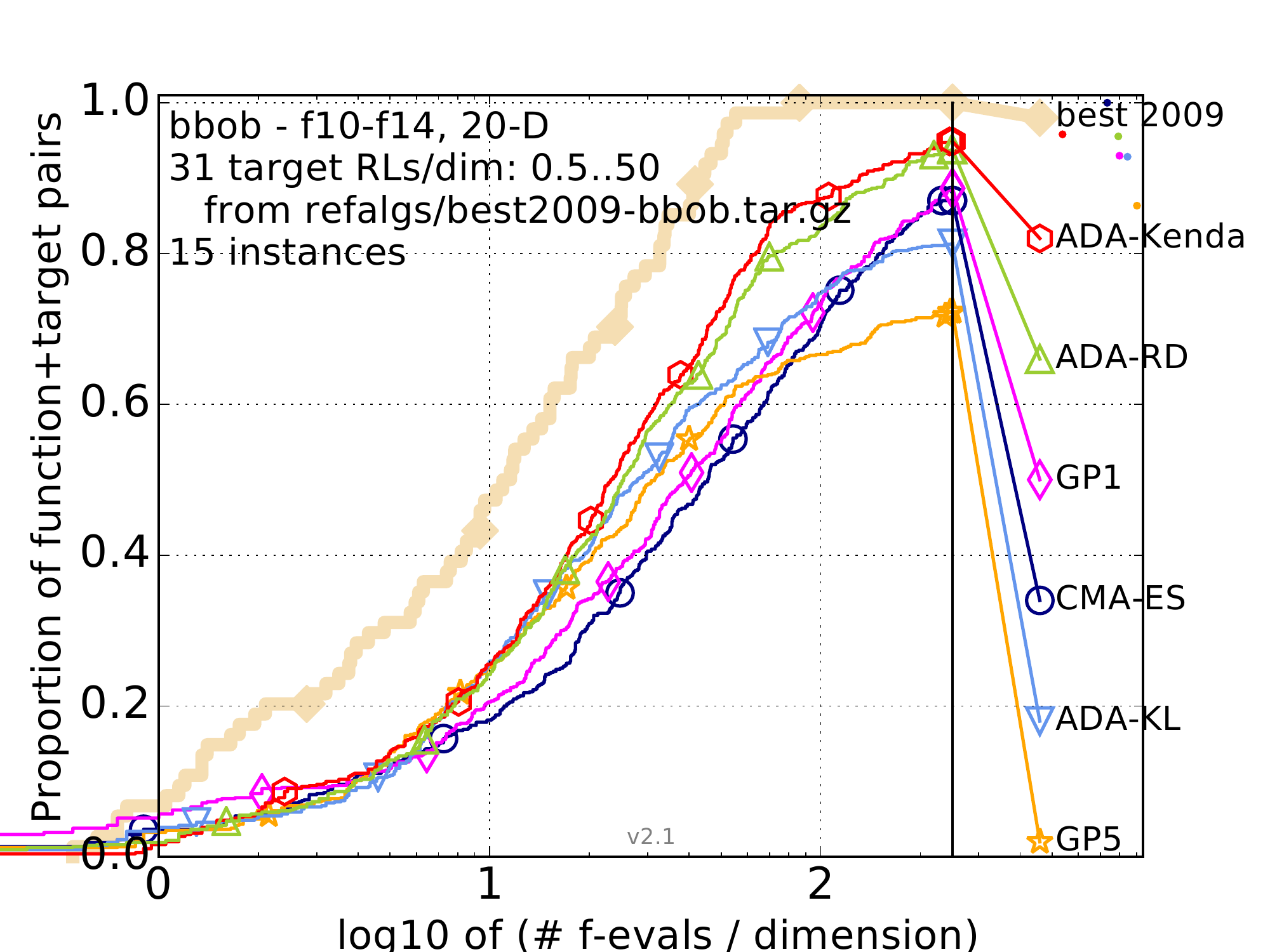} &
 \includegraphics[width=0.38\textwidth,trim=0mm 0mm 0mm 10mm, clip]{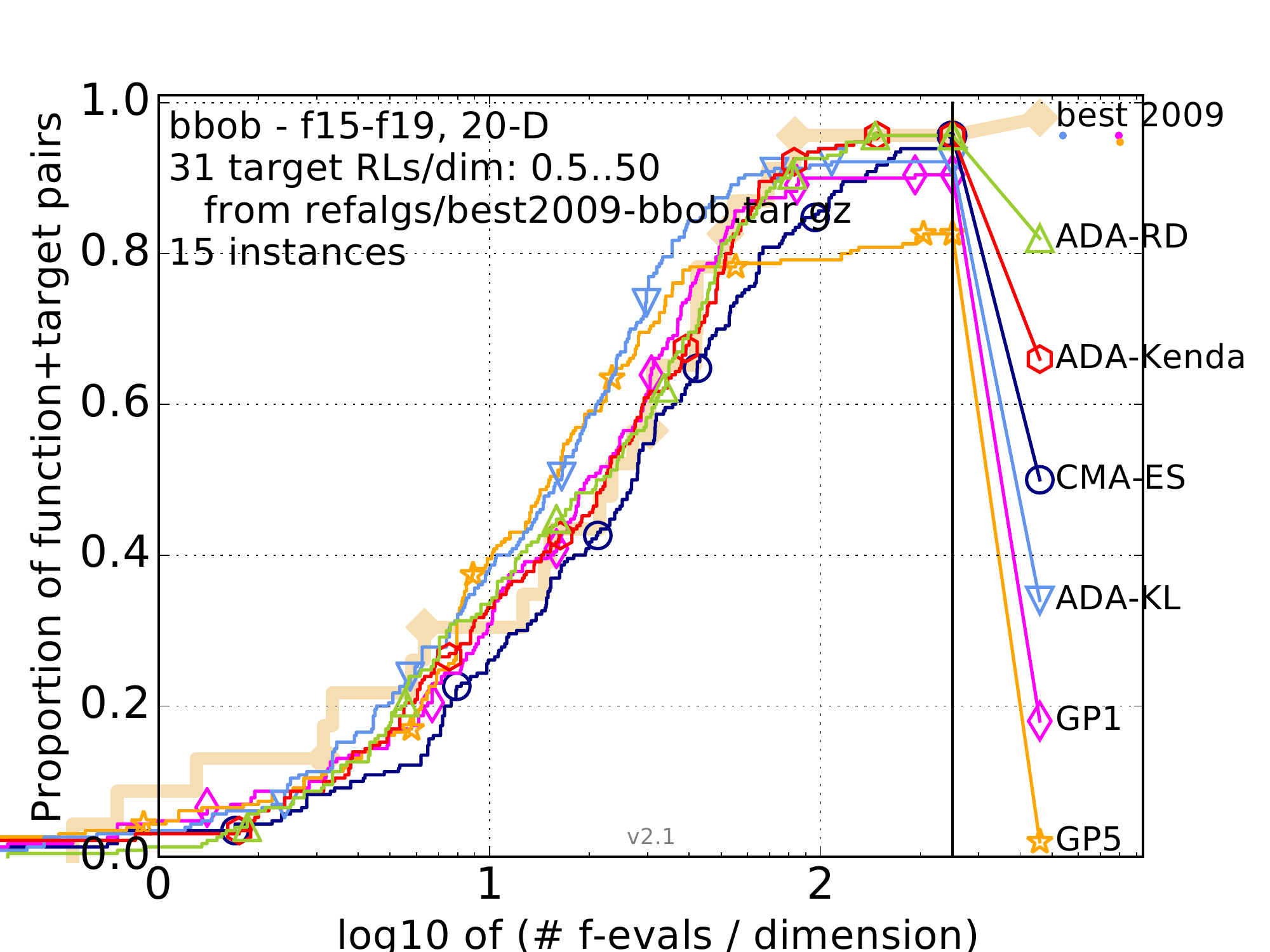} \\ 
 weakly structured multimodal functions & all functions\\
 \includegraphics[width=0.38\textwidth,trim=0mm 0mm 0mm 10mm, clip]{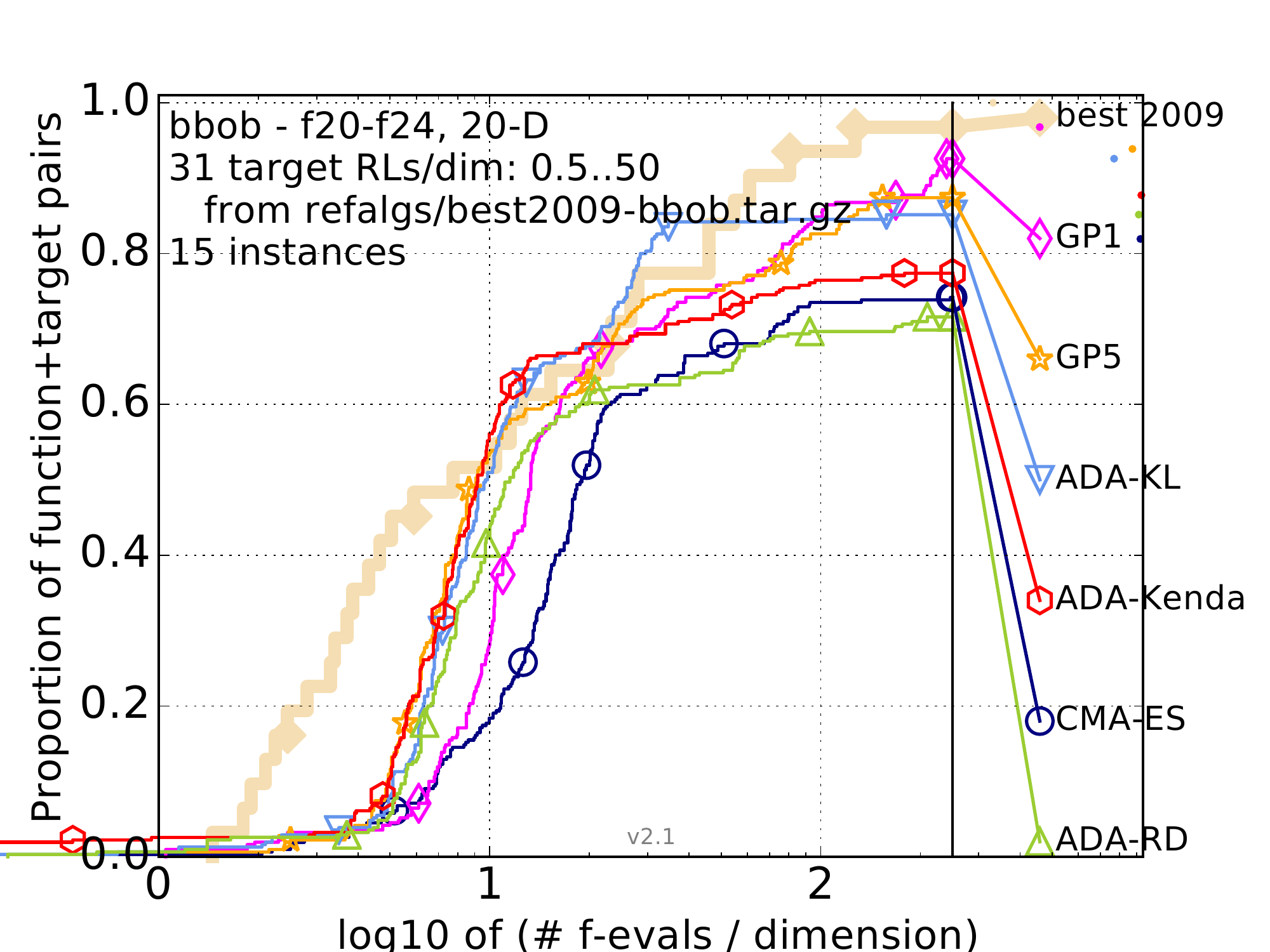} & 
 \includegraphics[width=0.38\textwidth,trim=0mm 0mm 0mm 10mm, clip]{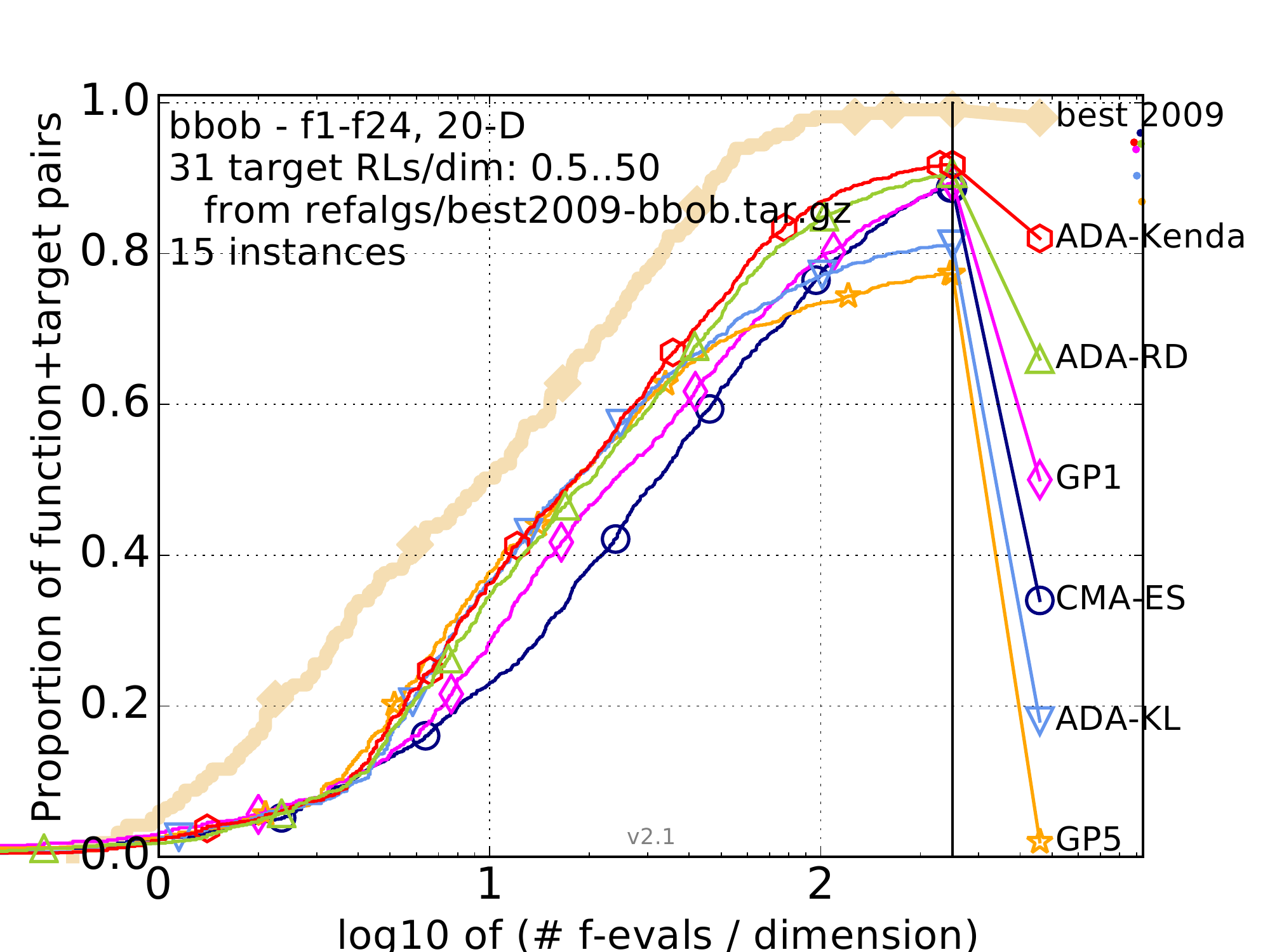} 
 \end{tabular}
\caption[\ecdfcaptiontwenty]{
  \ecdflabeltwenty
\bbobECDFslegend{20}
}
\end{figure*}

\section{Conclusion}
\label{sec:conclusion}
\normalsize
In this paper, we implemented several modifications of the Surrogate \cmaes\ (\scmaes), an algorithm using generation-based evolution control in connection with GPs.
We considered three measures of surrogate model error according to which an adequate number of upcoming model-evaluated generations could be estimated online.
Three resulting algorithms were compared on the COCO/BBOB framework with the \scmaes\ parametrized by two different numbers of consecutive model-evaluated generations.
Since the work on the adaptive extension is still in progress, the presented results summarize the performance of all compared algorithms on the whole BBOB framework or its function groups.
We found two error measures, the Kendall rank correlation and the rank difference error, that significantly outperformed the \scmaes\ used with a higher number of model-evaluated generations, especially in higher dimensionalities of the input space.
However, both of these algorithms provided only a minor improvement of the \scmaes\ used with a lower number of model-evaluated generations and in some tested scenarios fell behind both tested settings of the \scmaes.
An area for further research is the adjustment of other surrogate model parameters beside the control frequency, such as the number of the training points or the radius of the area from which they are selected.

\section{Acknowledgments}
The research reported in this paper has been supported by
the Czech Science Foundation (GAČR) grant 17-01251.

Access to computing and storage facilities owned by parties and projects contributing to the National Grid Infrastructure MetaCentrum, provided under the programme "Projects of Large Research, Development, and Innovations Infrastructures" (CESNET LM2015042), is greatly appreciated.

\bibliography{repicky2017adaptive}
\bibliographystyle{plain}
\end{document}

%% file: tables/adaTiming.tex
\begin{table}[h]
\centering
\caption[CPU timing for Adaptive \scmaes\ versions]{The time in seconds per function evaluation for the Adaptive \scmaes.}
\begin{tabular}{lrrrrr}
\toprule
Algorithm & \multicolumn{1}{c}{$2\dm$} & \multicolumn{1}{c}{$3\dm$} & \multicolumn{1}{c}{$5\dm$} & \multicolumn{1}{c}{$10\dm$} & \multicolumn{1}{c}{$20\dm$}\\
\midrule
ADA-KL      & $0.38$ & $0.26$ & $0.34$ & $0.69$ & $3.36$\\
ADA-Kendall & $0.47$ & $0.45$ & $0.61$ & $1.29$ & $6.27$\\
ADA-RD      & $0.57$ & $0.60$ & $0.71$ & $1.63$ & $7.90$\\
\bottomrule
\end{tabular}
\label{tab:timing:ada}
\end{table}

%% file: tables/adaStatsTable2.tex
\begin{table*}
\centering
\statstabcap
\begin{tabular}{ lrrrrrrrrrr }
\toprule
Dim& \multicolumn{2}{c}{$2\dm$}& \multicolumn{2}{c}{$3\dm$}& \multicolumn{2}{c}{$5\dm$}& \multicolumn{2}{c}{$10\dm$}& \multicolumn{2}{c}{$20\dm$}\\
\cmidrule(lr){1-1}
\cmidrule(lr){2-3}
\cmidrule(lr){4-5}
\cmidrule(lr){6-7}
\cmidrule(lr){8-9}
\cmidrule(lr){10-11}
{\large\sfrac{\nbFEs}{\bestFED}} & {\large\sfrac{1}{3}} & 1 & {\large\sfrac{1}{3}} & 1 & {\large\sfrac{1}{3}} & 1 & {\large\sfrac{1}{3}} & 1 & {\large\sfrac{1}{3}} & 1\\
\midrule
CMA-ES & $4.04$ & $4.25$ & $4.25$ & $3.96$ & $4.38$ & $3.58$ & $4.67$ & $3.83$ & $4.58$ & $4.42$\\
GP-1 & $3.38$ & $3.94$ & $4.21$ & $3.62$ & $3.92$ & $4.02$ & $3.69$ & $3.92$ & $3.54$ & $3.27$\\
GP-5 & $3.54$ & $3.12$ & $\mathbf{2.83}$ & $4.08$ & $3.81$ & $4.35$ & $4.25$ & $4.42$ & $4.23$ & $4.52$\\
ADA-KL & $3.23$ & $\mathbf{2.85}$ & $3.29$ & $3.44$ & $3.69$ & $3.73$ & $3.60$ & $4.04$ & $3.15$ & $3.65$\\
ADA-Ken & $3.98$ & $3.46$ & $3.25$ & $\mathbf{2.90}$ & $2.90$ & $2.96$ & $\mathbf{2.27}$ & $\mathbf{2.40}$ & $\mathbf{2.33}$ & $\mathbf{2.23}$\\
ADA-RD & $\mathbf{2.83}$ & $3.38$ & $3.17$ & $3.00$ & $\mathbf{2.31}$ & $\mathbf{2.35}$ & $2.52$ & $2.40$ & $3.17$ & $2.92$\\
\cmidrule(lr){1-1}
\cmidrule(lr){2-3}
\cmidrule(lr){4-5}
\cmidrule(lr){6-7}
\cmidrule(lr){8-9}
\cmidrule(lr){10-11}
$F_F$ & $1.48$ & $1.89$ & $2.52\statsig$ & $1.67$ & $4.47\statsig$ & $4.13\statsig$ & $7.82\statsig$ & $6.50\statsig$ & $5.35\statsig$ & $6.62\statsig$\\
\bottomrule
\end{tabular}
\statstablab
\end{table*}

%% file: tables/adaDuelTable2.tex
\begin{table*}
\dueltabcap
\setlength{\savetabcolsep}{\tabcolsep}
\setlength{\savecmidrulekern}{\cmidrulekern}

\setlength{\tabcolsep}{3pt}
\setlength{\cmidrulekern}{3pt}

\setlength{\dueltabcolw}{\textwidth-\firstcolw-26\tabcolsep}
\setlength{\dueltabcolw}{\dueltabcolw/12}

\settowidth{\astwidth}{${}^{\ast}$}
\centering
\newcolumntype{R}{>{\raggedleft\arraybackslash}m{\dueltabcolw}}
\begin{tabular}{ m{\firstcolw}RRRRRRRRRRRR }
\toprule
$\mathbf{2\textbf{D}}$ &\multicolumn{2}{c}{\parbox{2\dueltabcolw}{\centering\strut CMA-ES\strut}} & \multicolumn{2}{c}{\parbox{2\dueltabcolw}{\centering\strut GP-1\strut}} & \multicolumn{2}{c}{\parbox{2\dueltabcolw}{\centering\strut GP-5\strut}} & \multicolumn{2}{c}{\parbox{2\dueltabcolw}{\centering\strut ADA-KL\strut}} & \multicolumn{2}{c}{\parbox{2\dueltabcolw}{\centering\strut ADA-Ken\strut}} & \multicolumn{2}{c}{\parbox{2\dueltabcolw}{\centering\strut ADA-RD\strut}}\\
\cmidrule(lr){1-1}
\cmidrule(lr){2-3}
\cmidrule(lr){4-5}
\cmidrule(lr){6-7}
\cmidrule(lr){8-9}
\cmidrule(lr){10-11}
\cmidrule(lr){12-13}
{\large\sfrac{\nbFEs}{\bestFED}} & {\large\sfrac{1}{3}} & 1\mbox{\hspace{\astwidth}} & {\large\sfrac{1}{3}} & 1\mbox{\hspace{\astwidth}} & {\large\sfrac{1}{3}} & 1\mbox{\hspace{\astwidth}} & {\large\sfrac{1}{3}} & 1\mbox{\hspace{\astwidth}} & {\large\sfrac{1}{3}} & 1\mbox{\hspace{\astwidth}} & {\large\sfrac{1}{3}} & 1\mbox{\hspace{\astwidth}}\\
\midrule
CMA-ES & ---\makebox{\hspace{\astwidth}} & ---\makebox{\hspace{\astwidth}} &  8\makebox{\hspace{\astwidth}} &  8\makebox{\hspace{\astwidth}} &  11\makebox{\hspace{\astwidth}} &  8\makebox{\hspace{\astwidth}} &  10\makebox{\hspace{\astwidth}} &  7\makebox{\hspace{\astwidth}} &  11\makebox{\hspace{\astwidth}} &  10\makebox{\hspace{\astwidth}} &  7\makebox{\hspace{\astwidth}} &  9\makebox{\hspace{\astwidth}}\\
GP-1 &  16\makebox{\hspace{\astwidth}} &  16\makebox{\hspace{\astwidth}} & ---\makebox{\hspace{\astwidth}} & ---\makebox{\hspace{\astwidth}} &  12\makebox{\hspace{\astwidth}} &  9\makebox{\hspace{\astwidth}} &  13\makebox{\hspace{\astwidth}} &  8\makebox{\hspace{\astwidth}} &  13\makebox{\hspace{\astwidth}} &  9\makebox{\hspace{\astwidth}} &  9\makebox{\hspace{\astwidth}} &  6\makebox{\hspace{\astwidth}}\\
GP-5 &  13\makebox{\hspace{\astwidth}} &  16\makebox{\hspace{\astwidth}} &  12\makebox{\hspace{\astwidth}} &  15\makebox{\hspace{\astwidth}} & ---\makebox{\hspace{\astwidth}} & ---\makebox{\hspace{\astwidth}} &  11\makebox{\hspace{\astwidth}} &  10\makebox{\hspace{\astwidth}} &  14\makebox{\hspace{\astwidth}} &  13\makebox{\hspace{\astwidth}} &  9\makebox{\hspace{\astwidth}} &  14\makebox{\hspace{\astwidth}}\\
ADA-KL &  14\makebox{\hspace{\astwidth}} &  17\makebox{\hspace{\astwidth}} &  11\makebox{\hspace{\astwidth}} &  15\makebox{\hspace{\astwidth}} &  13\makebox{\hspace{\astwidth}} &  13\makebox{\hspace{\astwidth}} & ---\makebox{\hspace{\astwidth}} & ---\makebox{\hspace{\astwidth}} &  16\makebox{\hspace{\astwidth}} &  14\makebox{\hspace{\astwidth}} &  12\makebox{\hspace{\astwidth}} &  15\makebox{\hspace{\astwidth}}\\
ADA-Ken &  13\makebox{\hspace{\astwidth}} &  14\makebox{\hspace{\astwidth}} &  11\makebox{\hspace{\astwidth}} &  15\makebox{\hspace{\astwidth}} &  10\makebox{\hspace{\astwidth}} &  10\makebox{\hspace{\astwidth}} &  8\makebox{\hspace{\astwidth}} &  9\makebox{\hspace{\astwidth}} & ---\makebox{\hspace{\astwidth}} & ---\makebox{\hspace{\astwidth}} &  7\makebox{\hspace{\astwidth}} &  11\makebox{\hspace{\astwidth}}\\
ADA-RD &  17\makebox{\hspace{\astwidth}} &  15\makebox{\hspace{\astwidth}} &  15\makebox{\hspace{\astwidth}} &  17\makebox{\hspace{\astwidth}} &  15\makebox{\hspace{\astwidth}} &  9\makebox{\hspace{\astwidth}} &  12\makebox{\hspace{\astwidth}} &  9\makebox{\hspace{\astwidth}} &  17\makebox{\hspace{\astwidth}} &  12\makebox{\hspace{\astwidth}} & ---\makebox{\hspace{\astwidth}} & ---\makebox{\hspace{\astwidth}}\\
\midrule[\heavyrulewidth]
$\mathbf{3\textbf{D}}$ &\multicolumn{2}{c}{\parbox{2\dueltabcolw}{\centering\strut CMA-ES\strut}} & \multicolumn{2}{c}{\parbox{2\dueltabcolw}{\centering\strut GP-1\strut}} & \multicolumn{2}{c}{\parbox{2\dueltabcolw}{\centering\strut GP-5\strut}} & \multicolumn{2}{c}{\parbox{2\dueltabcolw}{\centering\strut ADA-KL\strut}} & \multicolumn{2}{c}{\parbox{2\dueltabcolw}{\centering\strut ADA-Ken\strut}} & \multicolumn{2}{c}{\parbox{2\dueltabcolw}{\centering\strut ADA-RD\strut}}\\
\cmidrule(lr){1-1}
\cmidrule(lr){2-3}
\cmidrule(lr){4-5}
\cmidrule(lr){6-7}
\cmidrule(lr){8-9}
\cmidrule(lr){10-11}
\cmidrule(lr){12-13}
{\large\sfrac{\nbFEs}{\bestFED}} & {\large\sfrac{1}{3}} & 1\mbox{\hspace{\astwidth}} & {\large\sfrac{1}{3}} & 1\mbox{\hspace{\astwidth}} & {\large\sfrac{1}{3}} & 1\mbox{\hspace{\astwidth}} & {\large\sfrac{1}{3}} & 1\mbox{\hspace{\astwidth}} & {\large\sfrac{1}{3}} & 1\mbox{\hspace{\astwidth}} & {\large\sfrac{1}{3}} & 1\mbox{\hspace{\astwidth}}\\
\midrule
CMA-ES & ---\makebox{\hspace{\astwidth}} & ---\makebox{\hspace{\astwidth}} &  11\makebox{\hspace{\astwidth}} &  9\makebox{\hspace{\astwidth}} &  7\makebox{\hspace{\astwidth}} &  13\makebox{\hspace{\astwidth}} &  8\makebox{\hspace{\astwidth}} &  10\makebox{\hspace{\astwidth}} &  9\makebox{\hspace{\astwidth}} &  9\makebox{\hspace{\astwidth}} &  7\makebox{\hspace{\astwidth}} &  8\makebox{\hspace{\astwidth}}\\
GP-1 &  13\makebox{\hspace{\astwidth}} &  15\makebox{\hspace{\astwidth}} & ---\makebox{\hspace{\astwidth}} & ---\makebox{\hspace{\astwidth}} &  7\makebox{\hspace{\astwidth}} &  14\makebox{\hspace{\astwidth}} &  7\makebox{\hspace{\astwidth}} &  11\makebox{\hspace{\astwidth}} &  6\makebox{\hspace{\astwidth}} &  8\makebox{\hspace{\astwidth}} &  10\makebox{\hspace{\astwidth}} &  9\makebox{\hspace{\astwidth}}\\
GP-5 &  17\makebox{\hspace{\astwidth}} &  11\makebox{\hspace{\astwidth}} &  17\makebox{\hspace{\astwidth}} &  10\makebox{\hspace{\astwidth}} & ---\makebox{\hspace{\astwidth}} & ---\makebox{\hspace{\astwidth}} &  15\makebox{\hspace{\astwidth}} &  9\makebox{\hspace{\astwidth}} &  13\makebox{\hspace{\astwidth}} &  6\makebox{\hspace{\astwidth}} &  14\makebox{\hspace{\astwidth}} &  9\makebox{\hspace{\astwidth}}\\
ADA-KL &  16\makebox{\hspace{\astwidth}} &  14\makebox{\hspace{\astwidth}} &  17\makebox{\hspace{\astwidth}} &  13\makebox{\hspace{\astwidth}} &  9\makebox{\hspace{\astwidth}} &  15\makebox{\hspace{\astwidth}} & ---\makebox{\hspace{\astwidth}} & ---\makebox{\hspace{\astwidth}} &  13\makebox{\hspace{\astwidth}} &  11\makebox{\hspace{\astwidth}} &  10\makebox{\hspace{\astwidth}} &  9\makebox{\hspace{\astwidth}}\\
ADA-Ken &  15\makebox{\hspace{\astwidth}} &  15\makebox{\hspace{\astwidth}} &  18\makebox{\hspace{\astwidth}} &  16\makebox{\hspace{\astwidth}} &  11\makebox{\hspace{\astwidth}} &  17\makebox{\hspace{\astwidth}} &  11\makebox{\hspace{\astwidth}} &  13\makebox{\hspace{\astwidth}} & ---\makebox{\hspace{\astwidth}} & ---\makebox{\hspace{\astwidth}} &  11\makebox{\hspace{\astwidth}} &  12\makebox{\hspace{\astwidth}}\\
ADA-RD &  17\makebox{\hspace{\astwidth}} &  16\makebox{\hspace{\astwidth}} &  14\makebox{\hspace{\astwidth}} &  15\makebox{\hspace{\astwidth}} &  10\makebox{\hspace{\astwidth}} &  14\makebox{\hspace{\astwidth}} &  14\makebox{\hspace{\astwidth}} &  15\makebox{\hspace{\astwidth}} &  13\makebox{\hspace{\astwidth}} &  10\makebox{\hspace{\astwidth}} & ---\makebox{\hspace{\astwidth}} & ---\makebox{\hspace{\astwidth}}\\
\midrule[\heavyrulewidth]
$\mathbf{5\textbf{D}}$ &\multicolumn{2}{c}{\parbox{2\dueltabcolw}{\centering\strut CMA-ES\strut}} & \multicolumn{2}{c}{\parbox{2\dueltabcolw}{\centering\strut GP-1\strut}} & \multicolumn{2}{c}{\parbox{2\dueltabcolw}{\centering\strut GP-5\strut}} & \multicolumn{2}{c}{\parbox{2\dueltabcolw}{\centering\strut ADA-KL\strut}} & \multicolumn{2}{c}{\parbox{2\dueltabcolw}{\centering\strut ADA-Ken\strut}} & \multicolumn{2}{c}{\parbox{2\dueltabcolw}{\centering\strut ADA-RD\strut}}\\
\cmidrule(lr){1-1}
\cmidrule(lr){2-3}
\cmidrule(lr){4-5}
\cmidrule(lr){6-7}
\cmidrule(lr){8-9}
\cmidrule(lr){10-11}
\cmidrule(lr){12-13}
{\large\sfrac{\nbFEs}{\bestFED}} & {\large\sfrac{1}{3}} & 1\mbox{\hspace{\astwidth}} & {\large\sfrac{1}{3}} & 1\mbox{\hspace{\astwidth}} & {\large\sfrac{1}{3}} & 1\mbox{\hspace{\astwidth}} & {\large\sfrac{1}{3}} & 1\mbox{\hspace{\astwidth}} & {\large\sfrac{1}{3}} & 1\mbox{\hspace{\astwidth}} & {\large\sfrac{1}{3}} & 1\mbox{\hspace{\astwidth}}\\
\midrule
CMA-ES & ---\makebox{\hspace{\astwidth}} & ---\makebox{\hspace{\astwidth}} &  8\makebox{\hspace{\astwidth}} &  12\makebox{\hspace{\astwidth}} &  11\makebox{\hspace{\astwidth}} &  14\makebox{\hspace{\astwidth}} &  11\makebox{\hspace{\astwidth}} &  14\makebox{\hspace{\astwidth}} &  7\makebox{\hspace{\astwidth}} &  10\makebox{\hspace{\astwidth}} &  2\makebox{\hspace{\astwidth}} &  8\makebox{\hspace{\astwidth}}\\
GP-1 &  16\makebox{\hspace{\astwidth}} &  12\makebox{\hspace{\astwidth}} & ---\makebox{\hspace{\astwidth}} & ---\makebox{\hspace{\astwidth}} &  11\makebox{\hspace{\astwidth}} &  12\makebox{\hspace{\astwidth}} &  11\makebox{\hspace{\astwidth}} &  12\makebox{\hspace{\astwidth}} &  9\makebox{\hspace{\astwidth}} &  7\makebox{\hspace{\astwidth}} &  3\makebox{\hspace{\astwidth}} &  4\makebox{\hspace{\astwidth}}\\
GP-5 &  13\makebox{\hspace{\astwidth}} &  10\makebox{\hspace{\astwidth}} &  13\makebox{\hspace{\astwidth}} &  12\makebox{\hspace{\astwidth}} & ---\makebox{\hspace{\astwidth}} & ---\makebox{\hspace{\astwidth}} &  10\makebox{\hspace{\astwidth}} &  6\makebox{\hspace{\astwidth}} &  9\makebox{\hspace{\astwidth}} &  5\makebox{\hspace{\astwidth}} &  8\makebox{\hspace{\astwidth}} &  7\makebox{\hspace{\astwidth}}\\
ADA-KL &  13\makebox{\hspace{\astwidth}} &  10\makebox{\hspace{\astwidth}} &  13\makebox{\hspace{\astwidth}} &  11\makebox{\hspace{\astwidth}} &  14\makebox{\hspace{\astwidth}} &  18\makebox{\hspace{\astwidth}} & ---\makebox{\hspace{\astwidth}} & ---\makebox{\hspace{\astwidth}} &  7\makebox{\hspace{\astwidth}} &  10\makebox{\hspace{\astwidth}} &  8\makebox{\hspace{\astwidth}} &  5\makebox{\hspace{\astwidth}}\\
ADA-Ken &  17\makebox{\hspace{\astwidth}} &  14\makebox{\hspace{\astwidth}} &  15\makebox{\hspace{\astwidth}} &  17\makebox{\hspace{\astwidth}} &  15\makebox{\hspace{\astwidth}} &  19\makebox{\hspace{\astwidth}} &  17\makebox{\hspace{\astwidth}} &  14\makebox{\hspace{\astwidth}} & ---\makebox{\hspace{\astwidth}} & ---\makebox{\hspace{\astwidth}} &  10\makebox{\hspace{\astwidth}} &  9\makebox{\hspace{\astwidth}}\\
ADA-RD &  22\makebox[0pt][l]{$^{\ast}$}\makebox{\hspace{\astwidth}} &  16\makebox{\hspace{\astwidth}} &  21\makebox[0pt][l]{$^{\ast}$}\makebox{\hspace{\astwidth}} &  20\makebox[0pt][l]{$^{\ast}$}\makebox{\hspace{\astwidth}} &  16\makebox[0pt][l]{$^{\ast}$}\makebox{\hspace{\astwidth}} &  17\makebox[0pt][l]{$^{\ast}$}\makebox{\hspace{\astwidth}} &  16\makebox{\hspace{\astwidth}} &  19\makebox{\hspace{\astwidth}} &  14\makebox{\hspace{\astwidth}} &  14\makebox{\hspace{\astwidth}} & ---\makebox{\hspace{\astwidth}} & ---\makebox{\hspace{\astwidth}}\\
\midrule[\heavyrulewidth]
$\mathbf{10\textbf{D}}$ &\multicolumn{2}{c}{\parbox{2\dueltabcolw}{\centering\strut CMA-ES\strut}} & \multicolumn{2}{c}{\parbox{2\dueltabcolw}{\centering\strut GP-1\strut}} & \multicolumn{2}{c}{\parbox{2\dueltabcolw}{\centering\strut GP-5\strut}} & \multicolumn{2}{c}{\parbox{2\dueltabcolw}{\centering\strut ADA-KL\strut}} & \multicolumn{2}{c}{\parbox{2\dueltabcolw}{\centering\strut ADA-Ken\strut}} & \multicolumn{2}{c}{\parbox{2\dueltabcolw}{\centering\strut ADA-RD\strut}}\\
\cmidrule(lr){1-1}
\cmidrule(lr){2-3}
\cmidrule(lr){4-5}
\cmidrule(lr){6-7}
\cmidrule(lr){8-9}
\cmidrule(lr){10-11}
\cmidrule(lr){12-13}
{\large\sfrac{\nbFEs}{\bestFED}} & {\large\sfrac{1}{3}} & 1\mbox{\hspace{\astwidth}} & {\large\sfrac{1}{3}} & 1\mbox{\hspace{\astwidth}} & {\large\sfrac{1}{3}} & 1\mbox{\hspace{\astwidth}} & {\large\sfrac{1}{3}} & 1\mbox{\hspace{\astwidth}} & {\large\sfrac{1}{3}} & 1\mbox{\hspace{\astwidth}} & {\large\sfrac{1}{3}} & 1\mbox{\hspace{\astwidth}}\\
\midrule
CMA-ES & ---\makebox{\hspace{\astwidth}} & ---\makebox{\hspace{\astwidth}} &  7\makebox{\hspace{\astwidth}} &  12\makebox{\hspace{\astwidth}} &  13\makebox{\hspace{\astwidth}} &  14\makebox{\hspace{\astwidth}} &  10\makebox{\hspace{\astwidth}} &  14\makebox{\hspace{\astwidth}} &  1\makebox{\hspace{\astwidth}} &  8\makebox{\hspace{\astwidth}} &  1\makebox{\hspace{\astwidth}} &  4\makebox{\hspace{\astwidth}}\\
GP-1 &  17\makebox{\hspace{\astwidth}} &  12\makebox{\hspace{\astwidth}} & ---\makebox{\hspace{\astwidth}} & ---\makebox{\hspace{\astwidth}} &  15\makebox{\hspace{\astwidth}} &  15\makebox{\hspace{\astwidth}} &  14\makebox{\hspace{\astwidth}} &  14\makebox{\hspace{\astwidth}} &  4\makebox{\hspace{\astwidth}} &  5\makebox{\hspace{\astwidth}} &  5\makebox{\hspace{\astwidth}} &  4\makebox{\hspace{\astwidth}}\\
GP-5 &  11\makebox{\hspace{\astwidth}} &  10\makebox{\hspace{\astwidth}} &  9\makebox{\hspace{\astwidth}} &  9\makebox{\hspace{\astwidth}} & ---\makebox{\hspace{\astwidth}} & ---\makebox{\hspace{\astwidth}} &  8\makebox{\hspace{\astwidth}} &  11\makebox{\hspace{\astwidth}} &  6\makebox{\hspace{\astwidth}} &  4\makebox{\hspace{\astwidth}} &  8\makebox{\hspace{\astwidth}} &  5\makebox{\hspace{\astwidth}}\\
ADA-KL &  14\makebox{\hspace{\astwidth}} &  10\makebox{\hspace{\astwidth}} &  10\makebox{\hspace{\astwidth}} &  10\makebox{\hspace{\astwidth}} &  16\makebox{\hspace{\astwidth}} &  13\makebox{\hspace{\astwidth}} & ---\makebox{\hspace{\astwidth}} & ---\makebox{\hspace{\astwidth}} &  8\makebox{\hspace{\astwidth}} &  5\makebox{\hspace{\astwidth}} &  9\makebox{\hspace{\astwidth}} &  8\makebox{\hspace{\astwidth}}\\
ADA-Ken &  23\makebox[0pt][l]{$^{\ast}$}\makebox{\hspace{\astwidth}} &  16\makebox[0pt][l]{$^{\ast}$}\makebox{\hspace{\astwidth}} &  20\makebox{\hspace{\astwidth}} &  19\makebox[0pt][l]{$^{\ast}$}\makebox{\hspace{\astwidth}} &  18\makebox[0pt][l]{$^{\ast}$}\makebox{\hspace{\astwidth}} &  20\makebox[0pt][l]{$^{\ast}$}\makebox{\hspace{\astwidth}} &  16\makebox{\hspace{\astwidth}} &  19\makebox[0pt][l]{$^{\ast}$}\makebox{\hspace{\astwidth}} & ---\makebox{\hspace{\astwidth}} & ---\makebox{\hspace{\astwidth}} &  13\makebox{\hspace{\astwidth}} &  12\makebox{\hspace{\astwidth}}\\
ADA-RD &  23\makebox[0pt][l]{$^{\ast}$}\makebox{\hspace{\astwidth}} &  20\makebox[0pt][l]{$^{\ast}$}\makebox{\hspace{\astwidth}} &  19\makebox{\hspace{\astwidth}} &  20\makebox[0pt][l]{$^{\ast}$}\makebox{\hspace{\astwidth}} &  16\makebox[0pt][l]{$^{\ast}$}\makebox{\hspace{\astwidth}} &  19\makebox[0pt][l]{$^{\ast}$}\makebox{\hspace{\astwidth}} &  15\makebox{\hspace{\astwidth}} &  15\makebox[0pt][l]{$^{\ast}$}\makebox{\hspace{\astwidth}} &  11\makebox{\hspace{\astwidth}} &  12\makebox{\hspace{\astwidth}} & ---\makebox{\hspace{\astwidth}} & ---\makebox{\hspace{\astwidth}}\\
\midrule[\heavyrulewidth]
$\mathbf{20\textbf{D}}$ &\multicolumn{2}{c}{\parbox{2\dueltabcolw}{\centering\strut CMA-ES\strut}} & \multicolumn{2}{c}{\parbox{2\dueltabcolw}{\centering\strut GP-1\strut}} & \multicolumn{2}{c}{\parbox{2\dueltabcolw}{\centering\strut GP-5\strut}} & \multicolumn{2}{c}{\parbox{2\dueltabcolw}{\centering\strut ADA-KL\strut}} & \multicolumn{2}{c}{\parbox{2\dueltabcolw}{\centering\strut ADA-Ken\strut}} & \multicolumn{2}{c}{\parbox{2\dueltabcolw}{\centering\strut ADA-RD\strut}}\\
\cmidrule(lr){1-1}
\cmidrule(lr){2-3}
\cmidrule(lr){4-5}
\cmidrule(lr){6-7}
\cmidrule(lr){8-9}
\cmidrule(lr){10-11}
\cmidrule(lr){12-13}
{\large\sfrac{\nbFEs}{\bestFED}} & {\large\sfrac{1}{3}} & 1\mbox{\hspace{\astwidth}} & {\large\sfrac{1}{3}} & 1\mbox{\hspace{\astwidth}} & {\large\sfrac{1}{3}} & 1\mbox{\hspace{\astwidth}} & {\large\sfrac{1}{3}} & 1\mbox{\hspace{\astwidth}} & {\large\sfrac{1}{3}} & 1\mbox{\hspace{\astwidth}} & {\large\sfrac{1}{3}} & 1\mbox{\hspace{\astwidth}}\\
\midrule
CMA-ES & ---\makebox{\hspace{\astwidth}} & ---\makebox{\hspace{\astwidth}} &  7\makebox{\hspace{\astwidth}} &  5\makebox{\hspace{\astwidth}} &  11\makebox{\hspace{\astwidth}} &  12\makebox{\hspace{\astwidth}} &  9\makebox{\hspace{\astwidth}} &  13\makebox{\hspace{\astwidth}} &  4\makebox{\hspace{\astwidth}} &  3\makebox{\hspace{\astwidth}} &  3\makebox{\hspace{\astwidth}} &  5\makebox{\hspace{\astwidth}}\\
GP-1 &  17\makebox{\hspace{\astwidth}} &  19\makebox{\hspace{\astwidth}} & ---\makebox{\hspace{\astwidth}} & ---\makebox{\hspace{\astwidth}} &  14\makebox{\hspace{\astwidth}} &  17\makebox{\hspace{\astwidth}} &  12\makebox{\hspace{\astwidth}} &  16\makebox{\hspace{\astwidth}} &  7\makebox{\hspace{\astwidth}} &  6\makebox{\hspace{\astwidth}} &  9\makebox{\hspace{\astwidth}} &  8\makebox{\hspace{\astwidth}}\\
GP-5 &  13\makebox{\hspace{\astwidth}} &  12\makebox{\hspace{\astwidth}} &  10\makebox{\hspace{\astwidth}} &  7\makebox{\hspace{\astwidth}} & ---\makebox{\hspace{\astwidth}} & ---\makebox{\hspace{\astwidth}} &  4\makebox{\hspace{\astwidth}} &  5\makebox{\hspace{\astwidth}} &  6\makebox{\hspace{\astwidth}} &  5\makebox{\hspace{\astwidth}} &  10\makebox{\hspace{\astwidth}} &  6\makebox{\hspace{\astwidth}}\\
ADA-KL &  15\makebox{\hspace{\astwidth}} &  11\makebox{\hspace{\astwidth}} &  12\makebox{\hspace{\astwidth}} &  8\makebox{\hspace{\astwidth}} &  20\makebox{\hspace{\astwidth}} &  19\makebox{\hspace{\astwidth}} & ---\makebox{\hspace{\astwidth}} & ---\makebox{\hspace{\astwidth}} &  9\makebox{\hspace{\astwidth}} &  8\makebox{\hspace{\astwidth}} &  13\makebox{\hspace{\astwidth}} &  10\makebox{\hspace{\astwidth}}\\
ADA-Ken &  20\makebox[0pt][l]{$^{\ast}$}\makebox{\hspace{\astwidth}} &  21\makebox[0pt][l]{$^{\ast}$}\makebox{\hspace{\astwidth}} &  17\makebox{\hspace{\astwidth}} &  18\makebox{\hspace{\astwidth}} &  18\makebox[0pt][l]{$^{\ast}$}\makebox{\hspace{\astwidth}} &  19\makebox[0pt][l]{$^{\ast}$}\makebox{\hspace{\astwidth}} &  15\makebox{\hspace{\astwidth}} &  16\makebox{\hspace{\astwidth}} & ---\makebox{\hspace{\astwidth}} & ---\makebox{\hspace{\astwidth}} &  17\makebox{\hspace{\astwidth}} &  17\makebox{\hspace{\astwidth}}\\
ADA-RD &  21\makebox{\hspace{\astwidth}} &  19\makebox[0pt][l]{$^{\ast}$}\makebox{\hspace{\astwidth}} &  15\makebox{\hspace{\astwidth}} &  16\makebox{\hspace{\astwidth}} &  14\makebox{\hspace{\astwidth}} &  18\makebox[0pt][l]{$^{\ast}$}\makebox{\hspace{\astwidth}} &  11\makebox{\hspace{\astwidth}} &  14\makebox{\hspace{\astwidth}} &  7\makebox{\hspace{\astwidth}} &  7\makebox{\hspace{\astwidth}} & ---\makebox{\hspace{\astwidth}} & ---\makebox{\hspace{\astwidth}}\\
\bottomrule
\end{tabular}

\setlength{\tabcolsep}{\savetabcolsep}
\setlength{\cmidrulekern}{\savecmidrulekern}
\dueltablab
\end{table*}

%% file: ppdata/cocopp_commands.tex
\newcommand{\algorithmA}{CMA-ES}
\newcommand{\algorithmB}{GP1}
\newcommand{\algorithmC}{GP5}
\newcommand{\algorithmD}{ADA-KL}
\newcommand{\algorithmE}{ADA-Kendall}
\newcommand{\algorithmF}{ADA-RD}
\newcommand{\bbobECDFslegend}[1]{
Bootstrapped empirical cumulative distribution of the number of objective function evaluations divided by dimension (FEvals/DIM) for all functions and subgroups in #1-D. The targets are chosen from $10^{[-8..2]}$ such that the best algorithm from BBOB 2009 just not reached them within a given budget of $k$ $\times$ DIM, with $31$ different values of $k$ chosen equidistant in logscale within the interval $\{0.5, \dots, 50\}$. The ``best 2009'' line corresponds to the best \aRT\ observed during BBOB 2009 for each selected target.
}
\newcommand{\bbobppfigslegend}[1]{
Legend: 
{\color{NavyBlue}$\circ$}:~\algorithmA
, {\color{Magenta}$\diamondsuit$}:~\algorithmB
, {\color{Orange}$\star$}:~\algorithmC
, {\color{CornflowerBlue}$\triangledown$}:~\algorithmD
, {\color{red}$\varhexagon$}:~\algorithmE
, {\color{YellowGreen}$\triangle$}:~\algorithmF
}
\definecolor{NavyBlue}{HTML}{000080}
\definecolor{Magenta}{HTML}{FF00FF}
\definecolor{Orange}{HTML}{FFA500}
\definecolor{CornflowerBlue}{HTML}{6495ED}
\definecolor{YellowGreen}{HTML}{9ACD32}
\definecolor{Gray}{HTML}{BEBEBE}
\definecolor{Yellow}{HTML}{FFFF00}
\definecolor{GreenYellow}{HTML}{ADFF2F}
\definecolor{ForestGreen}{HTML}{228B22}
\definecolor{Lavender}{HTML}{FFC0CB}
\definecolor{SkyBlue}{HTML}{87CEEB}
\definecolor{NavyBlue}{HTML}{000080}
\definecolor{Goldenrod}{HTML}{DDF700}
\definecolor{VioletRed}{HTML}{D02090}
\definecolor{CornflowerBlue}{HTML}{6495ED}
\definecolor{LimeGreen}{HTML}{32CD32}

\newcommand{\bbobpptablesmanylegend}[2]{%
        Average runtime (\aRT\ in number of function 
        evaluations) divided by the respective best \aRT\ measured during BBOB-2009 in
        #1.
        The \aRT\ and in braces, as dispersion measure, the half difference between 
        10 and 90\%-tile of bootstrapped run lengths appear for each algorithm and 
        run-length based target, the corresponding reference \aRT\
        (preceded by the target \Df-value in \textit{italics}) in the first row. 
        \#succ is the number of trials that reached the target value of the last column.
        The median number of conducted function evaluations is additionally given in 
        \textit{italics}, if the target in the last column was never reached. 
        Entries, succeeded by a star, are statistically significantly better (according to
        the rank-sum test) when compared to all other algorithms of the table, with
        $p = 0.05$ or $p = 10^{-k}$ when the number $k$ following the star is larger
        than 1, with Bonferroni correction of #2. A $\downarrow$ indicates the same tested against the best algorithm from BBOB 2009. Best results are printed in bold.
        \cocoversion}